\documentclass[sigconf]{acmart}

\renewcommand\footnotetextcopyrightpermission[1]{}
\AtBeginDocument{%
  }

\setcopyright{acmlicensed}
\copyrightyear{2026}
\acmYear{2026}
\acmDOI{XXXXXXX.XXXXXXX}
\acmConference[Under Review]{}
\acmISBN{}

\acmSubmissionID{2374}

\usepackage{amsmath}
\usepackage[ruled,linesnumbered]{algorithm2e}
\usepackage{tcolorbox}
\usepackage{colortbl}
\usepackage{xspace}
\usepackage{arydshln}
\usepackage[capitalize,noabbrev]{cleveref}
\usepackage{multirow}
\usepackage{booktabs,tabularx,makecell,array}
\newcolumntype{Y}{>{\centering\arraybackslash}X}

\renewcommand{\arraystretch}{1.10} 
\usepackage{xcolor}

\definecolor{sarback}{RGB}{247,241,236}   
\definecolor{sarframe}{RGB}{196,168,145}

\definecolor{ammback}{RGB}{245,243,235}   
\definecolor{ammframe}{RGB}{188,176,140}

\definecolor{mycolor_blue}{rgb}{0,0.2,0.8}
\definecolor{mycolor_green}{rgb}{0,0.5,0.1}
\definecolor{mycolor_pink}{rgb}{1.0,0.4,0.6}
\newcommand{\Paragraph}[1]{\noindent\textbf{#1}}

\def\eg{\textit{e.g.}\xspace}
\def\ie{\textit{i.e.}\xspace}

\def\vs{\textit{vs.}\xspace}

\newcommand{\subsecref}[1]{Section~\ref{subsec:#1}}
\newcommand{\figref}[1]{Fig.~\ref{fig:#1}}
\newcommand{\tabref}[1]{Table~\ref{tab:#1}}
\newcommand{\eqnref}[1]{Eq.~\eqref{eq:#1}}

\begin{document}

\title{FlowAnchor: Stabilizing the Editing Signal for Inversion-Free Video Editing}

\author{Ze Chen}
\authornote{Equal contribution.} 
\email{chenze@cuc.edu.cn}
\affiliation{%
  \institution{MIPG, Communication University of China}
  \city{Beijing}
  \country{China}
}

\author{Lan Chen}
\authornotemark[1]
\email{chenlaneva@mails.cuc.edu.cn}
\affiliation{%
  \institution{MIPG, Communication University of China}
  \city{Beijing}
  \country{China}
}

\author{Yuanhang Li}
\email{yuanhangli@cuc.edu.cn}
\affiliation{%
  \institution{MIPG, Communication University of China}
  \city{Beijing}
  \country{China}
}

\author{Qi Mao}
\authornote{Corresponding author.}
\email{qimao@cuc.edu.cn}
\affiliation{%
  \institution{MIPG, Communication University of China}
  \city{Beijing}
  \country{China}
}

\renewcommand{\shortauthors}{Chen et al.}

\begin{abstract}
We propose \textbf{FlowAnchor}, a training-free framework for stable and efficient inversion-free, flow-based video editing.
Inversion-free editing methods have recently shown impressive efficiency and structure preservation in images by directly steering the sampling trajectory with an editing signal.
However, extending this paradigm to videos remains challenging, often failing in multi-object scenes or with increased frame counts.
We identify the root cause as the \textit{instability of the editing signal} in high-dimensional video latent spaces, which arises from imprecise spatial localization and length-induced magnitude attenuation.
To overcome this challenge, FlowAnchor explicitly anchors both \emph{where} to edit and \emph{how strongly} to edit.
It introduces \textbf{Spatial-aware Attention Refinement}, which enforces consistent alignment between textual guidance and spatial regions, and \textbf{Adaptive Magnitude Modulation}, which adaptively preserves sufficient editing strength.
Together, these mechanisms stabilize the editing signal and guide the flow-based evolution toward the desired target distribution.
Extensive experiments demonstrate that \textbf{FlowAnchor} achieves more faithful, temporally coherent, and computationally efficient video editing across challenging multi-object and fast-motion scenarios.
The project page is available at \url{https://cuc-mipg.github.io/FlowAnchor.github.io/}.

\end{abstract}



\begin{teaserfigure}
  \includegraphics[width=0.99\textwidth]{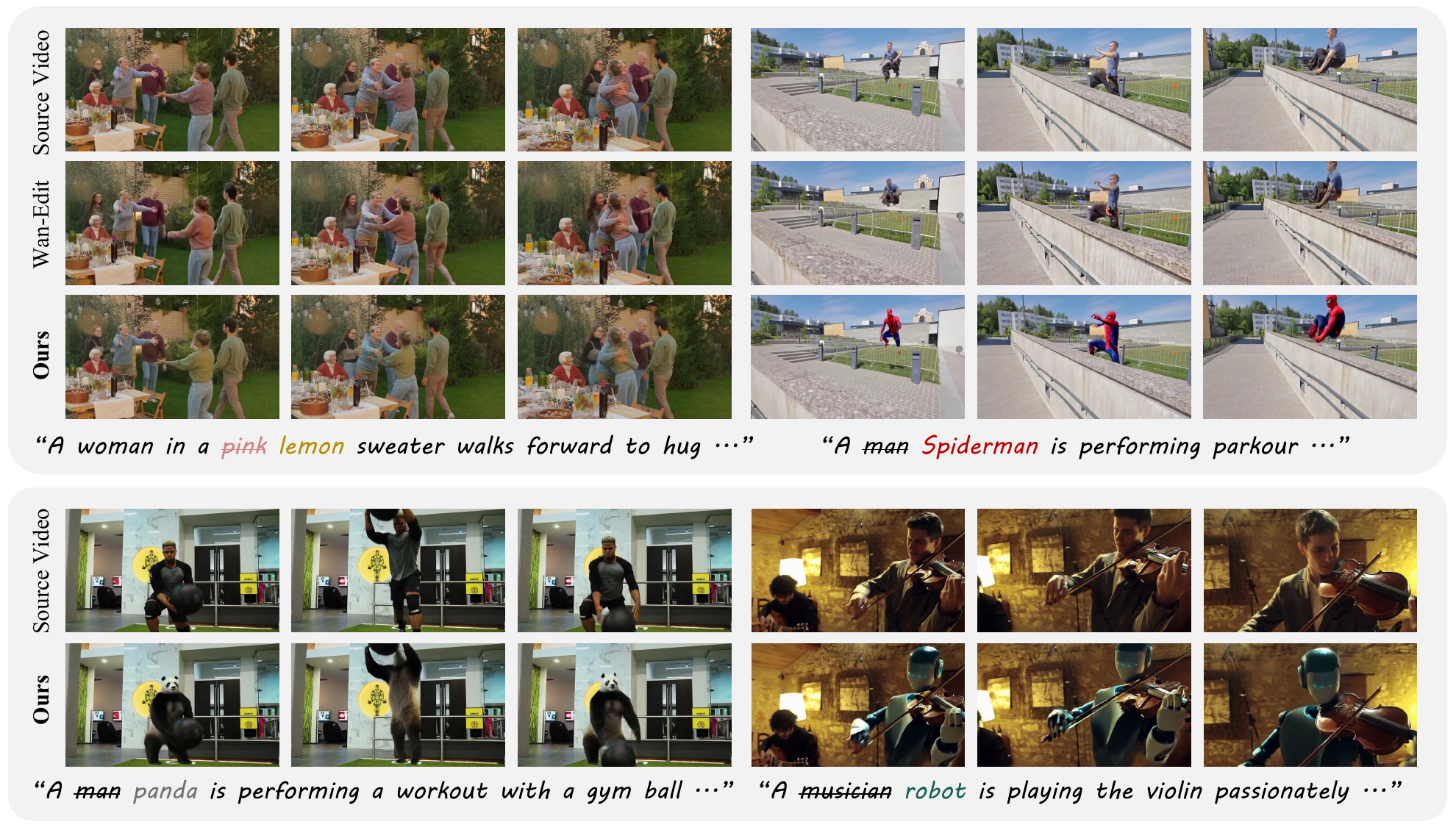}
  \vspace{-4 mm}
  \caption{\textbf{FlowAnchor stabilizes inversion-free video editing across diverse challenging scenarios.}
While the inversion-free baseline \textit{Wan-Edit}~\cite{li2025five} often struggles with mislocalized or weak edits, especially in multi-object scenes, fast-motion videos, and large semantic changes, our FlowAnchor achieves precise localized editing with improved temporal consistency, semantic faithfulness, and background preservation.
}
  \Description{Teaser figure showing representative video editing results across different scenes.}
  \label{fig:teaser}
  \vspace{2 mm}
\end{teaserfigure}


\maketitle

\section{Introduction}
\label{sec:intro}
Rapid and stable video editing is increasingly demanded in modern creative workflows, yet achieving high-fidelity edits with both speed and temporal stability remains an open challenge.
Previous approaches~\cite{wang2025taming, wang2025videodirector,geyertokenflow} predominantly rely on \textit{inversion}, which is computationally expensive and often introduces reconstruction errors that accumulate over time, leading to temporal drift.
While recent \textit{inversion-free} paradigms such as FlowEdit~\cite{kulikov2025flowedit} achieve fast, structure-preserving edits in images, the same idea breaks down in videos.
As shown in \figref{teaser}, naive adaptations such as Wan-Edit~\cite{li2025five} that treat videos as image batches fail in scenarios involving multiple objects or increased frame counts.

\begin{figure}[t]
    \centering
    \includegraphics[width=\columnwidth]{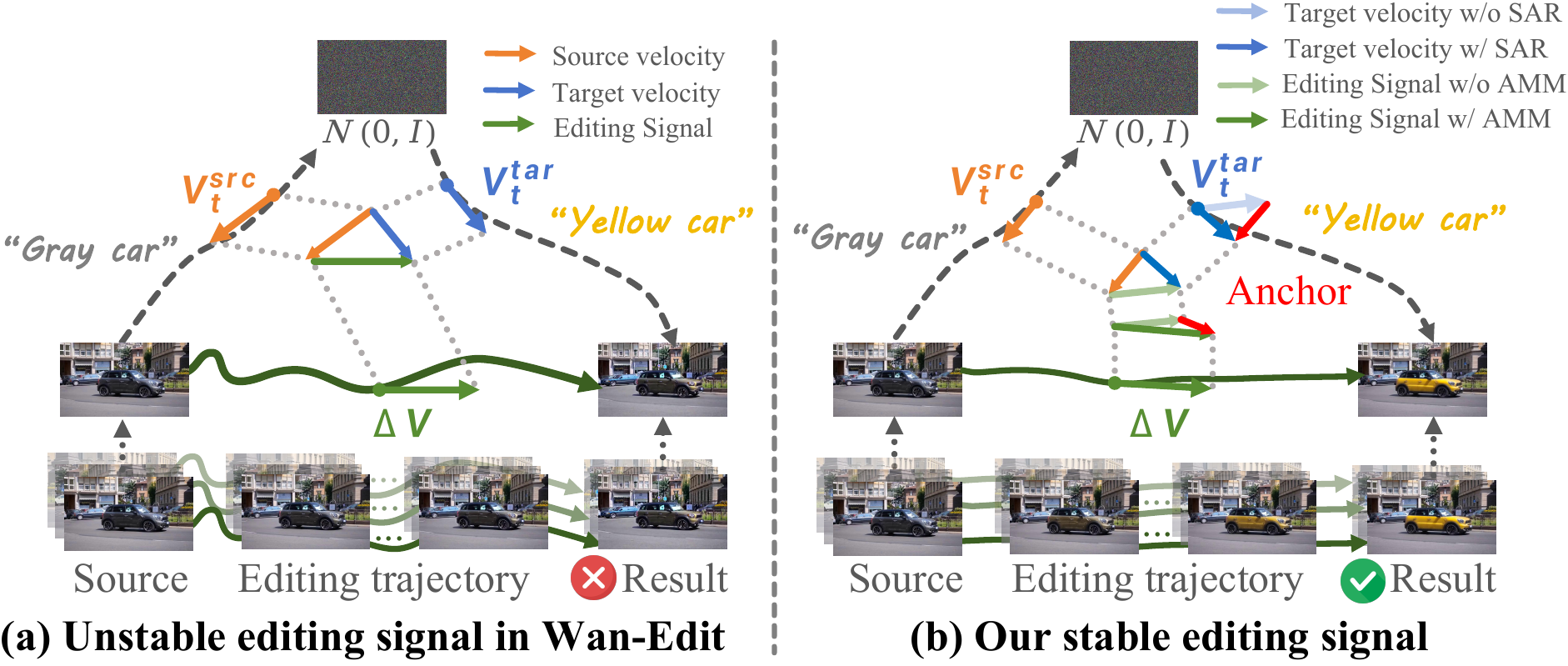}
    \vspace{-4 mm}
    \caption{\textbf{Wan-Edit~\cite{li2025five} vs.\ Ours.}
    (a) Naively extending FlowEdit~\cite{kulikov2025flowedit} to videos such as Wan-Edit~\cite{li2025five} produces unstable editing signals, causing the editing trajectory to distort and resulting in suboptimal edits. 
    (b) FlowAnchor provides an explicit anchor to stabilize the editing trajectory toward the intended target.}
    \vspace{-5 mm}
    \label{fig:anchor}
\end{figure}

We attribute this degradation to the \textbf{instability of the editing signal} in high-dimensional video latent spaces.
The editing signal, defined as the difference between source- and target-conditioned velocity fields in FlowEdit~\cite{kulikov2025flowedit}, is used to steer the editing trajectory from the source toward the target distribution.
Its instability manifests in two complementary aspects:
%
(1) \textbf{imprecise localization}, where the editing signal diffuses to irrelevant regions, leading to semantic misalignment;
and (2) \textbf{weakened magnitude}, where the signal attenuates as the temporal length increases.
Even when spatially localized, the signal may become too weak to effectively drive the latent trajectory toward the target distribution.
Together, these effects yield distorted evolution paths that deviate from the intended target video, as illustrated in \figref{anchor}(a).

To address this problem, we propose \textbf{FlowAnchor}, a training-free framework that stabilizes the editing signal by explicitly anchoring \emph{where} to edit and \emph{how strongly} to edit.
First, \textbf{Spatial-aware Attention Refinement} constrains cross-attention (CA) maps during velocity prediction with spatial priors, enforcing consistent alignment between textual guidance and the intended spatial regions.
This allows the editing signal to accurately capture the true regions of semantic variation.
Building on this precise localization, \textbf{Adaptive Magnitude Modulation} dynamically rescales the editing signal, ensuring sufficient strength to drive the edit.
As illustrated in \figref{anchor}(b), these two mechanisms jointly provide explicit anchors that rectify the editing signal, stabilizing the flow-based evolution toward the intended target distribution and yielding more faithful and temporally consistent edits as shown in~\figref{teaser}.

Our main contributions are summarized as follows:
\begin{itemize}
\item We are the first to identify and formalize \textit{editing-signal instability} as a key barrier in extending flow-based inversion-free editing to videos, and characterize two dominant failure modes: localization diffusion and length-induced magnitude attenuation.

\item We introduce \textbf{FlowAnchor}, a training-free framework that directly stabilizes the editing signal via \textbf{Spatial-aware Attention Refinement} for precise spatial localization and \textbf{Adaptive Magnitude Modulation} for robust strength.

\item We propose \textbf{Anchor-Bench}, a challenging benchmark featuring multi-object scenarios and fast-motion cases for evaluating localized video editing. We conduct extensive experiments on both FiVE-Bench and Anchor-Bench, demonstrating that our method consistently outperforms state-of-the-art baselines in text alignment, visual fidelity, temporal consistency, and computational efficiency.

\end{itemize}
\section{Related Work}
\label{sec:related}

\begin{figure*}[t]
    \centering
    \includegraphics[width=\textwidth]{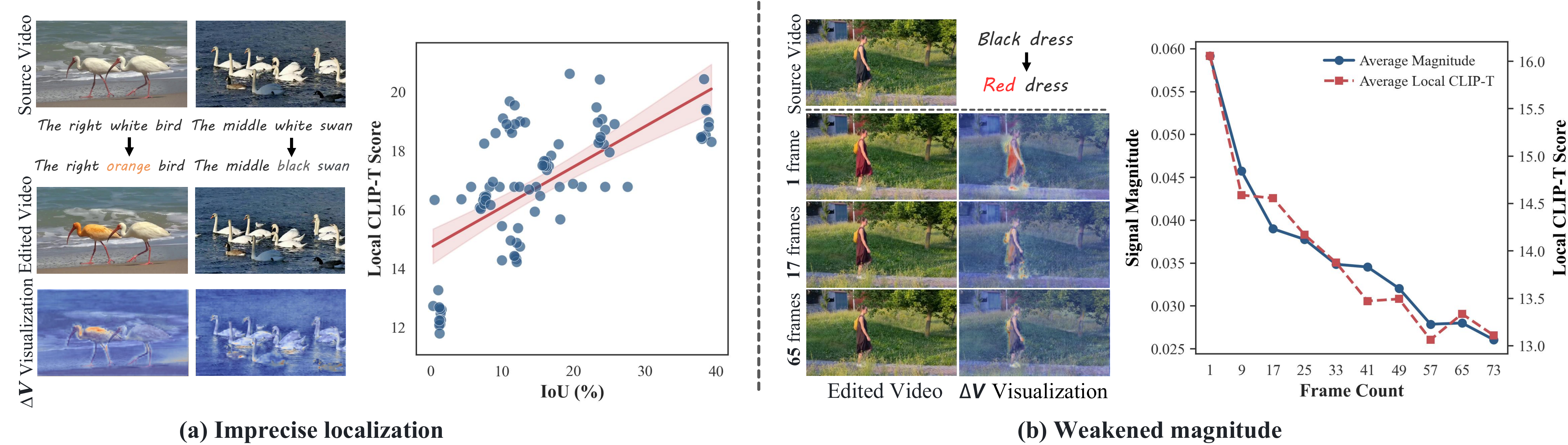}
    \vskip -0.1in
    \caption{\textbf{Challenges of unstable editing signals in existing inversion-free video editing.}
    (a) In multi-object scenes, the editing signal often fails to localize correctly, shifting to the wrong region or diffusing across the frame, leading to misplaced or ineffective editing.
    Statistically, the IoU between the binarized editing signal and the ground-truth mask varies widely across cases, and lower IoU correlates with lower Local CLIP-T, indicating weaker text--region alignment.
    (b) The magnitude of the editing signal drops noticeably as the number of frames increases, degrading editing effects even when spatial localization is correct.
    Both the signal magnitude and Local CLIP-T decrease with longer video length, showing weakened editing effects.}
    \label{fig:motivation}
\end{figure*}

\Paragraph{Video Diffusion Models.}
Early T2V models~\cite{ho2022video, singermake} adapt pretrained T2I architectures~\cite{rombach2022high} by inflating 2D U-Nets, successfully generating short video clips.
However, they struggle to preserve long-range temporal coherence due to limited temporal awareness.  
More recent developments in large-scale T2V diffusion models~\cite{yangcogvideox, kong2024hunyuanvideo, wan2025wan} adopt transformer-based architectures, notably the Diffusion Transformer (DiT)~\cite{peebles2023scalable}, which utilizes full 3D spatio-temporal attention to jointly model appearance, motion, and scene dynamics.  
This paradigm shift enables high-quality, temporally consistent video generation over long durations, laying a solid foundation for text-based video editing.

\Paragraph{Training-free Text-based Video Editing.}
Early text-based video editing methods~\cite{qi2023fatezero, ceylan2023pix2video, yang2023rerender, congflatten, geyertokenflow, zhang2023controlvideo, kara2024rave, yang2025videograin, wang2025videodirector} extend T2I diffusion models to the video domain, suffering from poor temporal coherence and noticeable flickering.
%
%
While methods like Rerender-A-Video~\cite{yang2023rerender}, FLATTEN~\cite{congflatten}, ControlVideo~\cite{zhang2023controlvideo}, and TokenFlow~\cite{geyertokenflow} make significant efforts to improve temporal consistency, they still struggle to achieve coherent results due to limitations of the backbone image generation model.
More recent work~\cite{wang2025taming, jiao2025uniedit, li2025five} leverages native T2V models~\cite{yangcogvideox, kong2024hunyuanvideo, wan2025wan} with learned spatio-temporal priors, showcasing improved temporal consistency and editing quality.
%

\Paragraph{Flow-based Video Editing.}
Flow-matching models~\cite{lipmanflow, liuflow} have emerged as powerful backbones for T2V generation.
Some methods~\cite{wang2025taming, jiao2025uniedit} follow the early inversion-based paradigm, which is computationally costly and prone to reconstruction errors.
Recently, inversion-free image editing methods~\cite{xu2024inversion, kulikov2025flowedit, yoonsplitflow, kim2025flowalign} bypass the inversion step, constructing a direct trajectory from source to target guided by the editing signal.
However, naively extending this paradigm to video~\cite{li2025five} often yields misaligned results.
While concurrent works attempt to improve quality, they either overlook the critical editing signal~\cite{cai2025dfvedit, li2025flowdirector} or rely on costly auxiliary conditions~\cite{kong2025taming}.
In contrast, we identify the editing signal as the primary bottleneck and propose effective solutions leveraging strong T2V priors.

\section{Method}
\label{sec:method}

\subsection{Preliminaries}

\Paragraph{Rectified Flow.}
Rectified Flow~\cite{lipmanflow, liuflow} defines a continuous transport map between two distributions $\pi_0$ and $\pi_1$ via an ordinary differential equation:
\begin{equation}
\frac{\mathrm{d}Z_t}{\mathrm{d}t} = V(Z_t, t), \quad t\in [0,1].
\label{eq:flow_ode}
\end{equation}
The marginal distribution at time $t$ is constrained to follow a linear interpolation between $X_0$ and $X_1$:
\begin{equation}
Z_t = (1-t)X_0 + tX_1,
\label{eq:linear_interp}
\end{equation}
which yields nearly straight trajectories for efficient and stable generation.
For text-conditioned models, the velocity field becomes $V(Z_t, t, C)$, where $C$ is the text condition.

\Paragraph{Inversion-free Editing with Rectified Flow.}
FlowEdit~\cite{kulikov2025flowedit} proposes an inversion-free method that constructs a direct path between the source distribution (guided by source prompt $\mathcal{P}$) and the target distribution (guided by target prompt $\mathcal{P}^*$).
Unlike inversion-based methods, it iteratively updates the \emph{editing trajectory} $Z_t^{\text{edit}}$ by estimating a velocity difference field $\Delta V_{t_i}$ to guide the transportation.
Specifically, at each step, a pseudo-source state $Z_{t_i}^{\text{src}}$ is sampled by linear interpolation between source image $X^{\text{src}}$ with noise $N_{t_i} \sim \mathcal{N}(0, I)$, coupled with a target state $Z_{t_i}$, defined as:
\begin{equation}
Z^{\text{tar}}_{t_i} = Z_{t_i}^{\text{edit}} + Z_{t_i}^{\text{src}} - X^{\text{src}}. 
\label{eq:z_tar}
\end{equation}
%
%
The editing trajectory then evolves by the \textbf{\emph{editing signal}} $\Delta V_{t_i}$:
\begin{equation}
\Delta V_{t_i} =
V(Z_{t_i}^{\text{tar}}, t_i, \mathcal{P}^*) -
V(Z_{t_i}^{\text{src}}, t_i, \mathcal{P}).
\label{eq:delta_v}
\end{equation}
Starting from the source image $X^{\text{src}}$, $Z_t^{\text{edit}}$ evolves as:
\begin{equation}
Z_{t_{i-1}}^{\text{edit}} = Z_{t_i}^{\text{edit}} + (t_{i-1} - t_i) \Delta V_{t_i},
\label{eq:flowedit_step}
\end{equation}
where $\Delta V_{t_i}$ represents the semantic discrepancy, guiding the editing trajectory.


\begin{figure*}[t]
    \centering
    \includegraphics[width=\textwidth]{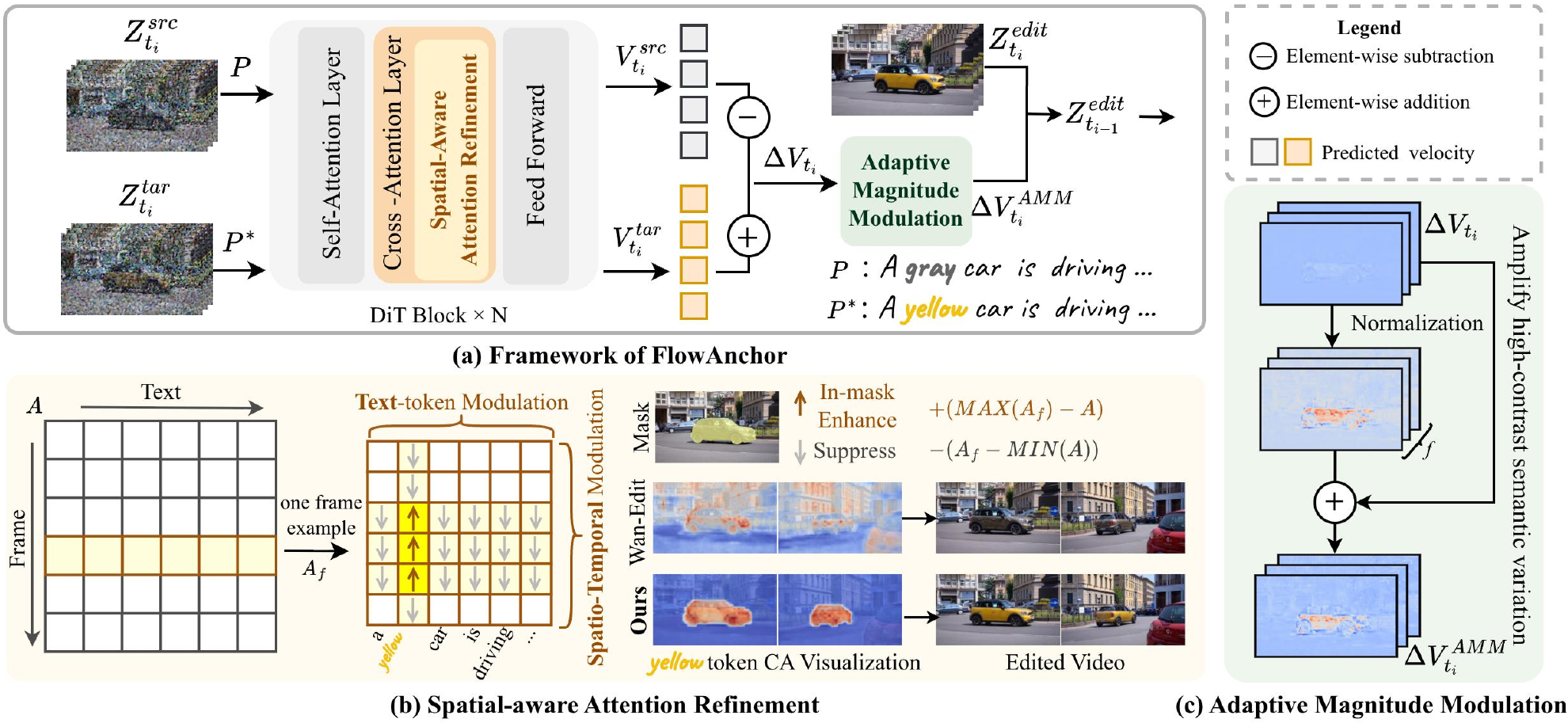}    
    \vskip -0.1in
    \caption{\textbf{Framework of FlowAnchor.}
    (a) At each timestep, $Z^{\text{src}}_{t_i}$ and $Z^{\text{tar}}_{t_i}$ are fed into the backbone model to obtain the corresponding velocity $V^{\text{src}}_{t_i}$ and $V^{\text{tar}}_{t_i}$.
    Within the backbone model, \textbf{SAR} injects a semantic alignment anchor, ensuring the editing signal $\Delta V_{t_i}= V^{\text{tar}}_{t_i} - V^{\text{src}}_{t_i}$ precisely captures the semantic variation in the target region.
    Then, \textbf{AMM} provides a magnitude anchor to enhance the semantic contrast.
    (b) The CA maps produced inside the backbone are modulated at the text token and spatio-temporal levels, enabling consistent localization associated with the target word across frames.
    After this modulation, the CA maps become strongly concentrated compared with Wan-Edit~\cite{li2025five}.
    (c) Once the editing signal $\Delta V_{t_i}$ is localized within the target region, it is further amplified by adding back its normalized map, further enhancing the semantic variation.}
    \label{fig:pipeline}
\end{figure*}

\subsection{Motivation}
\label{subsec:motivation}
While FlowEdit~\cite{kulikov2025flowedit} offers an efficient inversion-free framework, its naive application to video leads to noticeable performance degradation (\figref{teaser}).
%
We investigate this ineffectiveness by qualitatively and quantitatively analyzing the \emph{editing signal} $\Delta V$, revealing two factors that contribute to its instability.

\Paragraph{Imprecise Localization.} The editing signal often suffers from spatial misalignment, leading to semantic leakage in multi-object scenes.
For example, as shown in~\figref{motivation}(a), the signal may shift to an incorrect region, causing the ``orange'' effect to spill onto the other bird, or diffuse across the frame, losing focus on the intended ``black'' region.
This instability is quantitatively confirmed by the significant fluctuations in the IoU between the editing signal and ground-truth masks.
Moreover, a lower IoU correlates strongly with lower Local CLIP-T scores, indicating that imprecise spatial localization of the editing signal directly hinders the alignment between the target prompt and the editing region.

%

\Paragraph{Weakened Magnitude.} 
The editing signal magnitude diminishes as the video length increases.
As visualized in \figref{motivation}(b), the signal magnitude fades significantly in longer sequences, failing to drive the intended color change compared to the 1-frame baseline.
Statistical results further validate this trend: both the average signal magnitude and Local CLIP-T scores decay monotonically as the frame count rises, resulting in degraded performance.

\Paragraph{Analysis.}
Ideally, the target latent $Z^{\text{tar}}$ in \eqnref{z_tar} is coupled with the source latent $Z^{\text{src}}$ to enforce a shared noise field, as claimed in FlowEdit~\cite{kulikov2025flowedit}.
Given that the velocities $V(Z^{\text{tar}},t_i, \mathcal{P}^*)$ and $V(Z^{\text{src}}, t_i, \mathcal{P})$ are expected to remove approximately identical noise components, 
this formulation minimizes the transport cost between the source and target distributions, thereby theoretically guaranteeing maximal preservation of the original spatial structure.
%
%
%
However, in video generation, the injected source term inherently encodes strong spatiotemporal priors.
As the frame count increases, the spatiotemporal attention mechanism aggregates this dense source context over a larger number of frames.
%
Consequently, the dense source context outweighs the sparse editing semantics provided by the target prompt.
Consequently, the predicted target velocity $V(Z^{\text{tar}},t_i, \mathcal{P}^{\ast})$ becomes nearly identical to the source velocity $V(Z^{\text{src}},t_i, \mathcal{P})$, causing the editing signal $\Delta V$ to vanish.

\subsection{Spatial-aware Attention Refinement}
\label{subsec:spatial_anchoring}
As highlighted in \cref{subsec:motivation}, the editing signal suffers from imprecise localization.
We identify that this instability stems from the CA map, which governs the semantic alignment between the predicted velocities and the prompt but often lacks spatial precision in multi-object scenes.
%
To resolve this, we propose \textit{Spatial-aware Attention Refinement} (SAR), which provides the editing signal with a reliable spatial anchor by explicitly modulating the CA maps, as illustrated in~\figref{pipeline}(b).

Let $A \in \mathbb{R}^{(F \times H \times W)\times L}$ denote the CA maps. 
For an element $A_{i,j}$, $i \in \{1, \dots, F \times H \times W\}$ represents the spatio-temporal video token position, and $j \in \{1, \dots, L\}$ represents the text token index.
We define $J_{\text{tar}}$ as the index set of target text tokens driving the edit, and $M$ as the binary mask specifying the intended editing region.
To reinforce the correspondence between the target semantics ($J_{\text{tar}}$) and the localized visual region ($M$), SAR modulates the attention weights $A_{i,j}$ in two complementary steps, text-token modulation and spatio-temporal modulation.

\begin{table*}[!t]
\centering
\caption{\textbf{Quantitative comparisons on two benchmarks.} Warp-Err is reported in $10^{-3}$. Bold and underlined numbers denote the best and second-best results within each benchmark, respectively. $\dag$ denotes mask-based methods.}
\label{tab:baselines}
\vspace{-2mm}
\renewcommand{\arraystretch}{1.05}
\setlength{\tabcolsep}{2.0pt}
\small
\resizebox{\textwidth}{!}{%
\begin{tabular}{lcccccccccccc}
\toprule

\multirow{3}{*}{\textbf{Method}}
  & \multicolumn{6}{c}{\textbf{FiVE-Bench}}
  & \multicolumn{6}{c}{\textbf{Anchor-Bench}} \\
\cmidrule(lr){2-7}\cmidrule(lr){8-13}

  & \multicolumn{2}{c}{Text Alignment} & \multicolumn{2}{c}{Fidelity} & \multicolumn{2}{c}{Temporal Consistency}
  & \multicolumn{2}{c}{Text Alignment} & \multicolumn{2}{c}{Fidelity} & \multicolumn{2}{c}{Temporal Consistency} \\
\cmidrule(lr){2-3}\cmidrule(lr){4-5}\cmidrule(lr){6-7}\cmidrule(lr){8-9}\cmidrule(lr){10-11}\cmidrule(lr){12-13}

  & CLIP-T$\uparrow$
  & L.CLIP-T$\uparrow$
  & M.PSNR$\uparrow$
  & L.DINO$\uparrow$
  & CLIP-F$\uparrow$
  & Warp-Err$\downarrow$
  & CLIP-T$\uparrow$
  & L.CLIP-T$\uparrow$
  & M.PSNR$\uparrow$
  & L.DINO$\uparrow$
  & CLIP-F$\uparrow$
  & Warp-Err$\downarrow$ \\

\midrule

TokenFlow~\cite{geyertokenflow}
  & 28.22 & 19.84 & 18.43 & 0.7649 & 0.9630 & 3.378
  & 24.58 & 18.32 & 20.06 & 0.7790 & 0.9601 & 2.896 \\

VideoGrain$^\dag$~\cite{yang2025videograin}
  & 28.47 & \underline{21.07} & 26.03 & 0.7133 & 0.9397 & 4.705
  & 23.83 & \underline{20.47} & 24.49 & 0.7358 & 0.9340 & 4.079 \\

\midrule

RF-Solver~\cite{wang2025taming}
  & 27.92 & 19.68 & 20.02 & 0.7103 & 0.9561 & 7.703
  & 24.52 & 18.47 & 15.25 & 0.5913 & 0.9744 & 5.212 \\

UniEdit-Flow~\cite{jiao2025uniedit}
  & 27.95 & 20.39 & 23.96 & 0.6283 & 0.9563 & 4.539
  & 24.69 & 19.08 & 24.50 & 0.7814 & 0.9724 & \underline{1.695} \\

\midrule

Wan-Edit~\cite{li2025five}
  & 27.96 & 19.97 & 24.44 & 0.7346 & 0.9537 & 5.852
  & 23.07 & 18.43 & 25.18 & 0.8221 & 0.9712 & 2.578 \\

Wan-Edit+Mask$^\dag$
  & 26.85 & 19.79 & \underline{31.11} & \underline{0.7921} & 0.9574 & \underline{2.998}
  & 23.77 & 18.49 & \underline{26.63} & \underline{0.8423} & \underline{0.9748} & 2.173 \\

FlowDirector~\cite{li2025flowdirector}
  & \underline{28.61} & 20.44 & 22.81 & 0.5933 & \underline{0.9643} & 4.889
  & \textbf{25.19} & 19.57 & 19.93 & 0.5970 & 0.9720 & 2.030 \\

\rowcolor{gray!12}
\textbf{FlowAnchor (Ours)}$^\dag$
  & \textbf{28.82} & \textbf{21.50} & \textbf{31.18} & \textbf{0.8193} & \textbf{0.9703} & \textbf{2.386}
  & \underline{24.81} & \textbf{21.59} & \textbf{29.53} & \textbf{0.8504} & \textbf{0.9781} & \textbf{1.392} \\

\bottomrule
\end{tabular}%
}
\end{table*}
\Paragraph{Step 1: Text-Token Modulation.}
Within the edited region, we first strengthen the alignment with the target semantics by amplifying the CA map values of the target token while suppressing those of all other tokens.
For $i-th$ video token inside the mask ($M_i = 1$), we identify the current strongest and weakest attention responses across all $L$ text tokens:
\begin{equation}
\small
A^{\max}_{i} = \max\nolimits_{k \in \{1,\dots,L\}} A_{i,k}, \quad
A^{\min}_{i} = \min\nolimits_{k \in \{1,\dots,L\}} A_{i,k}.
\label{eq:token_bounds}
\end{equation}
We then modulate the attention map $A$ to $A'$ by pulling target tokens toward the maximum and suppressing non-target tokens toward the minimum:
\begin{equation}
\small
A'_{i,j} = \begin{cases}
A_{i,j} + \beta_1 \big(A^{\max}_{i} - A_{i,j}\big), & \text{if } M_i = 1,\; j \in J_{\text{tar}},\\[2pt]
A_{i,j} - \beta_1 \big(A_{i,j} - A^{\min}_{i}\big), & \text{if } M_i = 1,\; j \notin J_{\text{tar}},\\[2pt]
A_{i,j}, & \text{otherwise},
\end{cases}
\label{eq:token_mod}
\end{equation}
where $\beta_1 \in [0,1]$ controls the modulation strength.
This formulation increases the contrast between target and non-target responses, thereby sharpening the semantic focus of the editing signal.

\Paragraph{Step 2: Spatio-Temporal Modulation.}
While the text-token modulation in Step 1 effectively distinguishes target semantics, it ignores global temporal coherence. 
This leads to unstable attention distributions across consecutive frames, manifesting as flickering. 
To enforce spatio-temporal consistency, we regulate the cross-attention weights of the target tokens in $J_{\text{tar}}$ across the entire video sequence.
%
For each target token $j \in J_{\text{tar}}$, we first compute its maximum and minimum responses across all $F \times H \times W$ video tokens:
\begin{equation}
\small
A'^{\max}_{j} = \max\nolimits_{p \in \{1,\dots,F \times H \times W\}} A'_{p,j}, \quad
A'^{\min}_{j} = \min\nolimits_{p \in \{1,\dots,F \times H \times W\}} A'_{p,j}.
\label{eq:st_bounds}
\end{equation}
The attention map is then refined to $A''$:
\begin{equation}
\small
A''_{i,j} = \begin{cases}
A'_{i,j} + \beta_2 \big(A'^{\max}_{j} - A'_{i,j}\big),
& \text{if } M_i = 1,\; j \in J_{\text{tar}},\\[2pt]
A'_{i,j} - \beta_2 \big(A'_{i,j} - A'^{\min}_{j}\big),
& \text{if } M_i = 0,\; j \in J_{\text{tar}},\\[2pt]
A'_{i,j}, & \text{otherwise},
\end{cases}
\label{eq:st_mod}
\end{equation}
where $\beta_2 \in [0,1]$ controls the spatio-temporal modulation strength.
%
Jointly, Steps 1 and 2 provide an explicit anchor for ``where to edit''.
The resulting editing signal $\Delta V_{t_i}$ accurately captures the semantic variation within the target region, serving as a reliable foundation for the subsequent magnitude anchoring.

\begin{figure}[t]
    \centering
    \includegraphics[width=0.98\columnwidth]{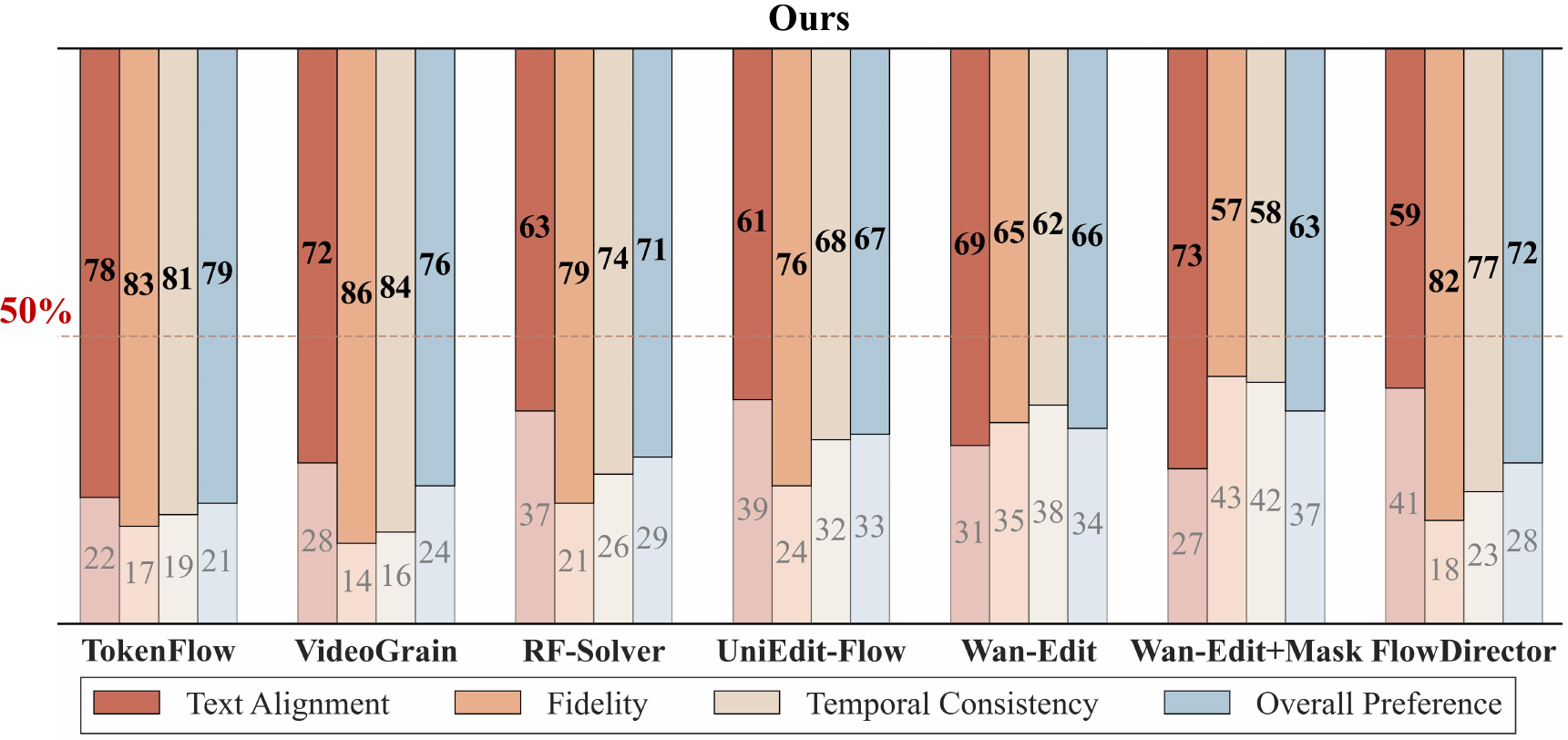}
    \vskip -0.05in
    \caption{\textbf{User preference study.} We report the preference rate (\%) of FlowAnchor (Ours) against each baseline across four aspects.
    FlowAnchor is consistently preferred over all baselines.}
    \vskip -0.05in
    \label{fig:user_study}
\end{figure}

\subsection{Adaptive Magnitude Modulation}
\label{subsec:amplitude_anchoring}
With the editing signal spatially anchored, we now address the second failure mode: the weakened magnitude that causes the editing signal to become insufficient for driving the trajectory toward the target distribution.
As analyzed in \cref{subsec:motivation}, the signal magnitude decays as the frame count $F$ increases.
%
To resolve this problem, we propose \textit{Adaptive Magnitude Modulation} (AMM), which exploits the intrinsic contrast of the editing signal to adaptively reinforce its magnitude, with a frame-aware scaling that directly compensates for the length-induced attenuation, as illustrated in~\figref{pipeline}(c).

\begin{figure*}[!t]
    \centering
    \includegraphics[width=\textwidth]{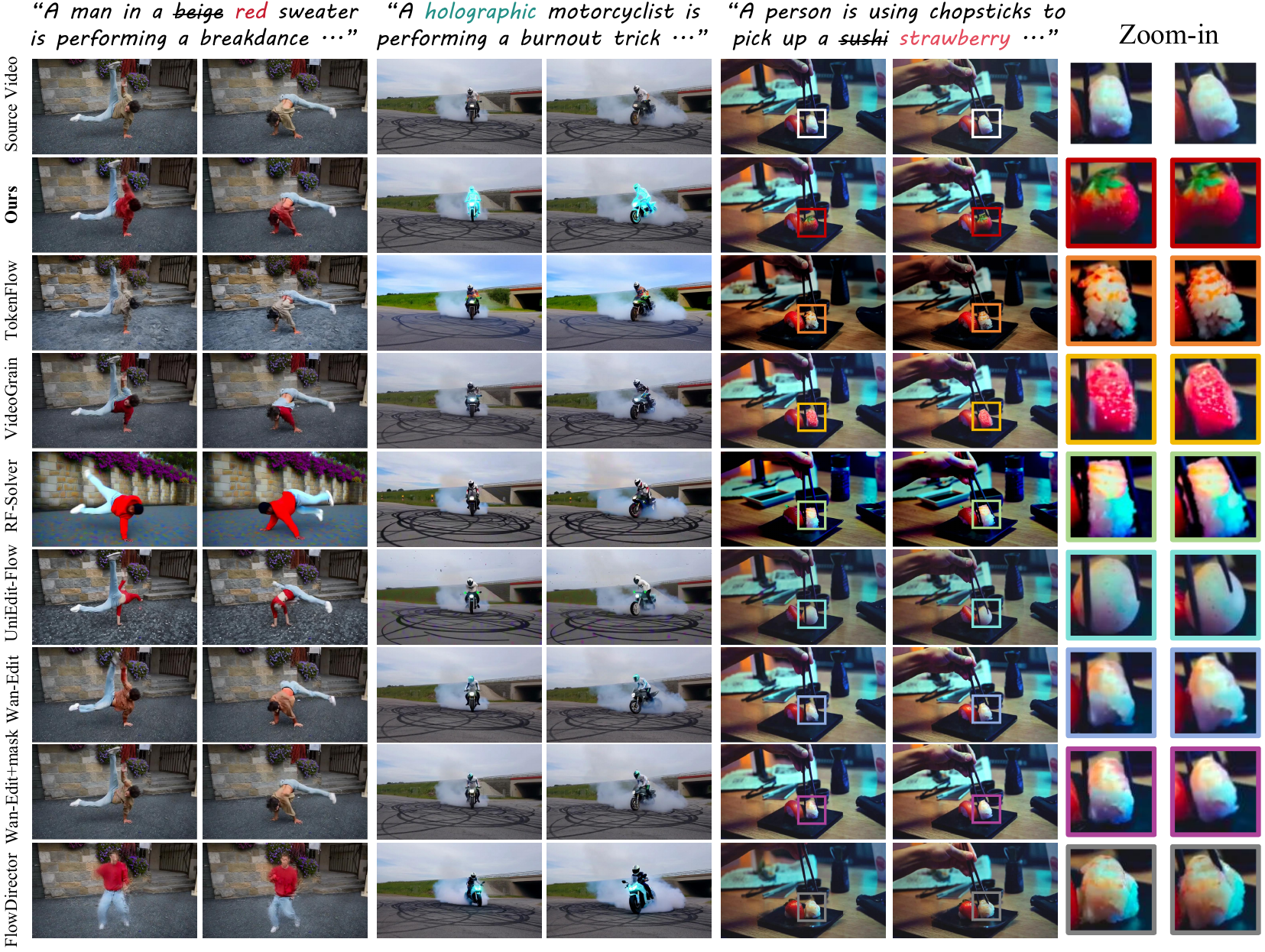}    
    \vskip -0.08in
    \caption{\textbf{Qualitative comparisons with baselines.}
    Our \textbf{FlowAnchor} outperforms baseline methods in both editing localization and effect quality, as well as in temporal consistency. \textbf{Zoom in for the best view of fine-grained details.}
    Please refer to the supplementary material for more results.}
    \label{fig:baseline}
\end{figure*}

Instead of applying a uniform global amplification that might inadvertently magnify background noise, our modulation adaptively reinforces the signal based on the signal's intrinsic intensity.
Building upon the precisely localized signal from the SAR, we use $\Delta V_{t_i}$ as an internal cue.
At each sampling step $t_i$, we first derive a contrast map $\mathcal{C}_{t_i}$ by applying max-min normalization to the editing signal:
\begin{equation}
\mathcal{C}_{t_i} = \frac{\Delta V_{t_i} - \min(\Delta V_{t_i})}{\max(\Delta V_{t_i}) - \min(\Delta V_{t_i})},
\label{eq:contrast_map}
\end{equation}
which assigns values near $1$ to regions with strong semantic variation and values near $0$ to background regions.
This map serves as a soft importance mask that identifies where the signal carries meaningful editing semantics.

Since our analysis demonstrates that an increased frame count attenuates the editing signal, we introduce a frame-adaptive amplification factor that provides monotonically increasing reinforcement for longer videos:
\begin{equation}
\gamma_F = \gamma \cdot \frac{\log F}{\log F_0},
\label{eq:gamma_f}
\end{equation}
where $\gamma > 0$ is the base amplification strength and $F_0$ is the model's default maximum length.
This design has two desirable properties.
First, it anchors the base amplification $\gamma$ at $F_0$. 
From this baseline, $\gamma_F$ adaptively increases with the actual frame count $F$, directly counteracting the intensified signal attenuation in longer sequences.
Second, when $F = 1$ (single-image editing), $\gamma_F = 0$, meaning no amplification is applied---this is consistent with the observation that FlowEdit~\cite{kulikov2025flowedit} already performs well in the image domain, where the editing signal does not suffer from length-induced attenuation.

The contrast map $\mathcal{C}_{t_i}$ is then combined with $\gamma_F$ to selectively reinforce the editing signal:
\begin{equation}
\Delta V_{t_i}^{\text{AMM}} = \big(1 + \gamma_F \cdot \mathcal{C}_{t_i}\big) \odot \Delta V_{t_i},
\label{eq:amp_anchor}
\end{equation}
where $\odot$ denotes element-wise multiplication.
Through this operation, entries of $\Delta V_{t_i}$ at high-contrast positions are amplified by a factor of up to $1+\gamma_F$, while background regions with near-zero contrast remain essentially unchanged.
The frame-adaptive factor $\gamma_F$ ensures that the reinforcement strength scales with the severity of magnitude attenuation: longer videos receive proportionally stronger compensation, directly addressing the length-induced signal weakening identified in \cref{subsec:motivation}.
Finally, the anchored editing signal drives the trajectory evolution:
\begin{equation}
Z^{\text{edit}}_{t_{i-1}}
=
Z^{\text{edit}}_{t_i}
+
\big(t_{i-1} - t_i\big)\,\Delta V_{t_i}^{\text{AMM}}.
\label{eq:anchored_step}
\end{equation}

\section{Experiments}
\label{sec:exp}

\subsection{Implementation Details}
Our method is built upon the widely-used DiT-based video generation model, Wan2.1-T2V-1.3B~\cite{wan2025wan}.
Following FlowEdit~\cite{kulikov2025flowedit}, we set inference steps $T{=}25$ and skip the first two steps to preserve the source layout.
SAR modulates all CA layers during $t \in [T, \tau]$ with $\tau{=}0.6T$, $\beta_1{=}\beta_2{=}0.3$; AMM is applied at every step with $\gamma{=}1.0$ and $F_0{=}21$.
\textit{\textbf{Notably, SAR is robust to mask quality, seamlessly accommodating precise masks from off-the-shelf segmenters, coarse bounding boxes, and hand-drawn scribbles.}} Further analysis is provided in \subsecref{mask} and Supplementary Materials (SM).
%
All experiments are conducted on one NVIDIA A800 GPU.

\subsection{Comparisons with Baselines}
\Paragraph{Datasets and Baselines.}
We evaluate on two benchmarks.
\textbf{(1) FiVE-Bench}~\cite{li2025five} contains 419 text-video editing pairs with precise masks, spanning object replacement (rigid \& non-rigid), addition, removal, color, and material editing, covering both real and generated videos.
However, it is largely limited to single-object scenes with most videos sourced from DAVIS~\cite{perazzi2016benchmark}.
\textbf{(2) Anchor-Bench} is our proposed benchmark comprising $74$ editing pairs of challenging multi-object real-world videos collected from the Internet, with up to $81$ frames at 480p resolution.
It covers color, material, and object replacement (rigid \& non-rigid) editing.
Prompts are generated by GPT-5 followed by manual refinement.
More details are provided in the SM.
We conduct a comprehensive comparison against seven state-of-the-art methods across three representative categories:
\textbf{(1) T2I-based methods}: TokenFlow~\cite{geyertokenflow} and VideoGrain~\cite{yang2025videograin};
\textbf{(2) inversion-based flow methods}: RF-Solver-Edit~\cite{wang2025taming} and UniEdit-Flow~\cite{jiao2025uniedit};
and \textbf{(3) inversion-free flow methods}: Wan-Edit~\cite{li2025five} and FlowDirector~\cite{li2025flowdirector}.
We also implement Wan-Edit+Mask by integrating masks into Wan-Edit.
Since VideoGrain~\cite{yang2025videograin} relies on precise spatial masks, \textbf{we generate masks for all mask-based methods on Anchor-Bench using SAM~\cite{kirillov2023segment}, ensuring a fair comparison.}

\Paragraph{Evaluation Metrics.}
We quantitatively compare all methods across four aspects.
\textbf{(1) Text Alignment.} We report CLIP-T~\cite{radford2021learning} to measure the global correspondence between the edited video and the entire target prompt and Local CLIP-T (L.CLIP-T), measuring localized semantic accuracy between the cropped region and the target words. 
\textbf{(2) Fidelity.} We use the masked PSNR (M.PSNR)~\cite{huynh2008scope} to measure the pixel-level reconstruction outside the mask, and Local DINO Similarity (L.DINO)~\cite{oquab2024dinov2} to measure the structure preservation between the edited videos and the source videos within the mask.
\textbf{(3) Temporal Consistency} is measured by CLIP-F~\cite{li2025flowdirector,yang2025videograin}, which assesses inter-frame semantic continuity, and Warp-Err~\cite{lai2018learning}, which quantifies pixel-level deviations via optical flow.
\textbf{(4) Efficiency.} We report the average inference time and peak GPU memory to measure computational efficiency under identical hardware conditions. 
\begin{table}[!t]\centering
\caption{\textbf{Ablation on SAR and AMM modules.} Warp-Err is reported in $10^{-3}$. TTM and STM denote Text-Token and Spatio-Temporal Modulation, respectively.}
\label{tab:ablation_module}
\vspace{-2mm}
\renewcommand{\arraystretch}{1.05}
\setlength{\tabcolsep}{6pt} 
\small
\begin{tabular}{lcccc}
\toprule
\textbf{Metric} & \textbf{w/o TTM} & \textbf{w/o STM} & \textbf{w/o AMM} & \textbf{Ours} \\
\midrule
CLIP-T$\uparrow$     & 24.38  & \underline{24.52}  & 22.65  & \textbf{24.81} \\
L.CLIP-T$\uparrow$   & 20.42  & \underline{20.86}  & 18.64  & \textbf{21.59} \\
M.PSNR$\uparrow$     & \underline{29.59} & 29.33  & \textbf{30.75} & 29.53 \\
L.DINO$\uparrow$     & \underline{0.8587} & 0.8349 & \textbf{0.9004} & 0.8504 \\
CLIP-F$\uparrow$     & \underline{0.9748} & 0.9742 & 0.9738 & \textbf{0.9781} \\
Warp-Err$\downarrow$ & 1.438  & 1.425  & \textbf{1.026} & \underline{1.392} \\
\bottomrule
\end{tabular}
\vspace{-1 mm}
\end{table}
\begin{figure}[!t]
    \centering
    \includegraphics[width=0.98\columnwidth]{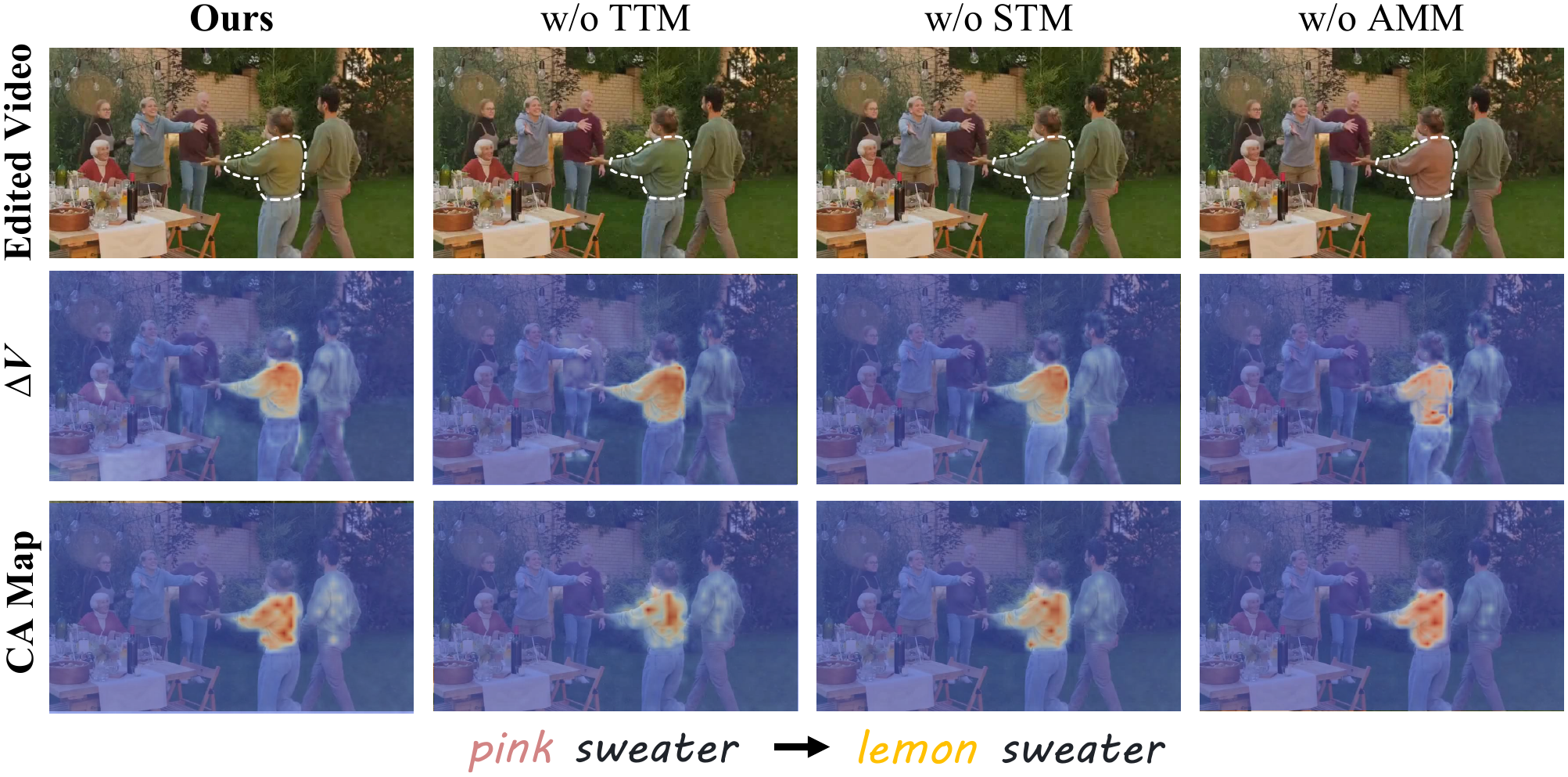}
    \vskip -0.05in
    \caption{\textbf{Qualitative analysis on SAR and AMM modules.}
    Both TTM and STM in SAR contribute to localizing the editing signal via cross-attention alignment, while AMM amplifies it for sufficient strength. Jointly, they ensure precise editing.
}
    \vskip -0.05in
    \label{fig:ablation}
\end{figure}

%

\Paragraph{Quantitative Results.}
We quantitatively evaluate our method against baselines using both automatic metrics and human evaluations.

\begin{figure}[!t]
    \centering
    \includegraphics[width=0.98\columnwidth]{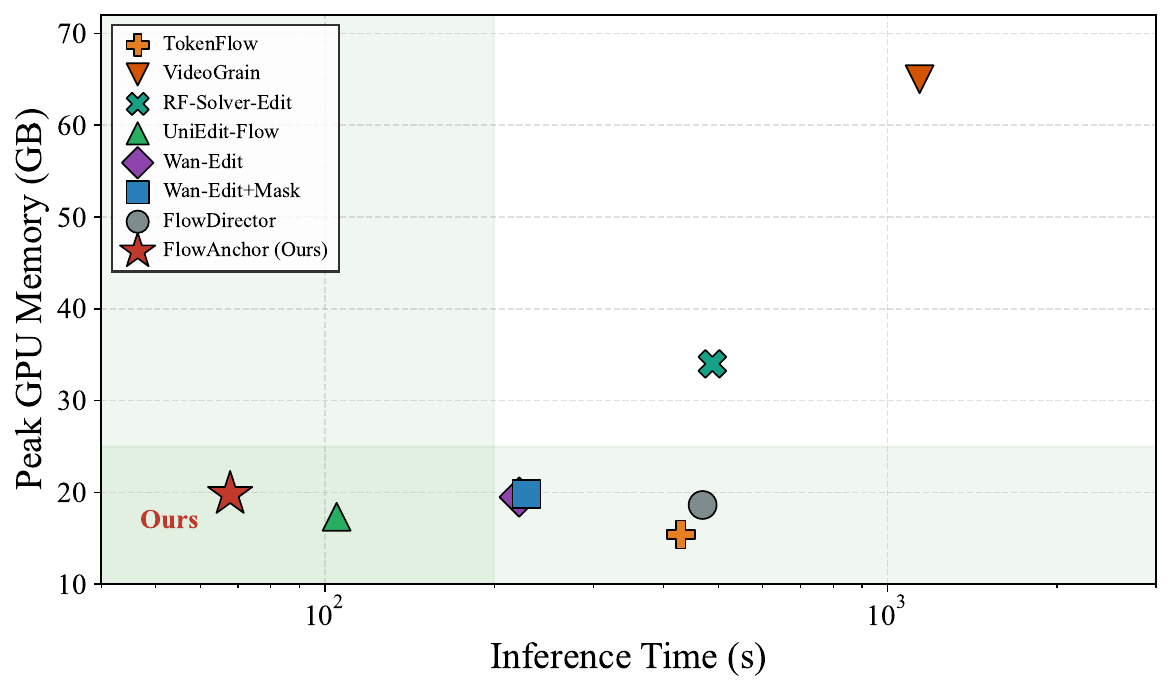}
    \vskip -0.1in
    \caption{\textbf{Efficiency comparison across methods.} Our method achieves the lowest inference time while maintaining competitive GPU memory usage, demonstrating a favorable trade-off between efficiency and editing quality.}
    \vspace{-3 mm}
    \label{fig:efficiency}
\end{figure}

\textit{Automatic Metrics.}
As shown in \tabref{baselines} and \figref{efficiency}, our method achieves the highest L.CLIP-T score on both FiVE-Bench and Anchor-Bench, demonstrating superior localized alignment with the target prompt.
Simultaneously, it maintains strong source fidelity and temporal coherence, as evidenced by the best M.PSNR, L.DINO, and CLIP-F scores.
Furthermore, our method achieves the lowest inference time among all baselines, highlighting its practical efficiency.

\textit{User Study.}
We conduct a user preference study with 20 participants through pairwise comparisons, evaluating text alignment, fidelity, temporal consistency, and overall preference.
Across all aspects, our method is consistently favored over the baselines, as shown in \figref{user_study}.

\Paragraph{Qualitative Results.}
\figref{baseline} presents qualitative comparisons across baselines.
For text alignment, TokenFlow~\cite{geyertokenflow}, Wan-Edit~\cite{li2025five} and Wan-Edit+Mask exhibit ineffective editing across multiple cases, \eg, failing to produce the ``red'' sweater in the breakdance case or the ``holographic'' effect in the motorcyclist case.
For fidelity, both RF-Solver-Edit~\cite{wang2025taming} and UniEdit-Flow~\cite{jiao2025uniedit} suffer from severe reconstruction errors in the breakdance case, with distorted human appearances and altered backgrounds.
FlowDirector~\cite{li2025flowdirector} also fails to preserve the structure of the human subject in this case.
In the ``strawberry'' case, RF-Solver-Edit and FlowDirector mislocalize the editing signal to incorrect regions, producing visible artifacts.
For temporal coherence, VideoGrain~\cite{yang2025videograin} exhibits noticeable flickering in the breakdance case, with inconsistent ``red'' appearances across frames.
In contrast, our method achieves accurate text alignment, preserves fidelity in both edited regions and the background, and maintains temporal consistency even under fast and large motions.

\begin{table}[t]\centering
\caption{\textbf{Hyperparameter sensitivity analysis.} Warp-Err is reported in $10^{-3}$. Each group varies one factor while keeping others at default: $\beta_1{=}0.3,\,\beta_2{=}0.3,\,\gamma{=}1.0,\,\tau{=}0.6T$.}\label{tab:hyperparameter}
\vspace{-2mm}
\renewcommand{\arraystretch}{1.05}
\setlength{\tabcolsep}{4pt}
\small
\resizebox{\columnwidth}{!}{%
\begin{tabular}{lcc cc cc c}
\toprule
\multirow{2}{*}{\textbf{Metric}}
  & \multicolumn{2}{c}{\textbf{SAR} $(\beta_1,\beta_2)$}
  & \multicolumn{2}{c}{\textbf{AMM} $\gamma$}
  & \multicolumn{2}{c}{\textbf{Timestep} $\tau$}
  & \multirow{2}{*}{\textbf{Ours}} \\
\cmidrule(lr){2-3}\cmidrule(lr){4-5}\cmidrule(lr){6-7}
  & (0.1,\,0.1)
  & (0.5,\,0.5)
  & 0.5
  & 1.5
  & 0.8$T$
  & 0.4$T$
  & \\
\midrule
CLIP-T$\uparrow$     & 23.81 & 24.02 & 22.61 & 23.59 & 23.82 & \underline{24.65} & \textbf{24.81} \\
L.CLIP-T$\uparrow$   & 18.91 & 19.65 & 18.77 & 19.96 & 19.24 & \underline{21.32} & \textbf{21.59} \\
M.PSNR$\uparrow$     & 29.15 & 29.02 & \textbf{30.80} & 25.59 & 29.16 & 29.24 & \underline{29.53} \\
L.DINO$\uparrow$     & \underline{0.8508} & 0.8109 & \textbf{0.9146} & 0.7027 & 0.8504 & 0.8235 & 0.8504 \\
CLIP-F$\uparrow$     & 0.9744 & 0.9739 & 0.9740 & 0.9689 & 0.9742 & 0.9740 & \textbf{0.9781} \\
Warp-Err$\downarrow$ & \underline{0.987} & 1.009 & \textbf{0.938} & 1.005 & 1.456 & 1.487 & 1.392 \\
\bottomrule
\end{tabular}%
}
\vspace{-3mm}
\end{table}

\subsection{Ablation Study}

\Paragraph{Impact of SAR.} 
As shown in \figref{ablation} and \tabref{ablation_module}, disabling either TTM ($\beta_1=0$) or STM ($\beta_2=0$) leads to imprecise localization and degraded CLIP-T scores, confirming the necessity of both constraints.
%
%
%
%
Regarding modulation strength, results in \tabref{hyperparameter} indicate that low values (\eg $0.1$) fail to provide sufficient guidance, while excessive values (\eg $0.5$) tend to degrade the fidelity.
Consequently, we select $\beta_1=\beta_2=0.3$ for the optimal balance.
For qualitative results, please refer to the SM.

\Paragraph{Impact of AMM.} Removing AMM (\ie $\gamma=0$) substantially reduces signal magnitude, leading to negligible editing effects (\figref{ablation}) and dropped CLIP-T scores (\tabref{ablation_module}).
Further analysis (\tabref{hyperparameter}) on $\gamma \in \{0.5, 1.0, 1.5\}$ indicates a trade-off: low values ($0.5$) result in insufficient strength as evidenced by low CLIP-T scores, while excessive values ($1.5$) cause structural distortion, indicated by a drop in L.DINO.
%
%
Thus, we adopt $\gamma = 1.0$ for the optimal balance.

\Paragraph{Impact of SAR Application Timesteps.}
\tabref{hyperparameter} reveals a clear trade-off regarding the application window $[T, \tau]$ of SAR.
A premature termination (\ie, $\tau=0.8T$) weakens text alignment (lower CLIP-T and L.CLIP-T), showing that the editing signal needs sufficient steps to establish a precise spatial anchor. 
However, overextending SAR to $\tau=0.4T$ harms fidelity and temporal consistency (lower M.PSNR, L.DINO, and higher Warp-Err). 
%
%
%
%
Therefore, we adopt $\tau=0.6T$ as the final configuration.
For qualitative results, please refer to the SM.


\begin{figure}[!t]
    \centering
    \includegraphics[width=0.97\linewidth]{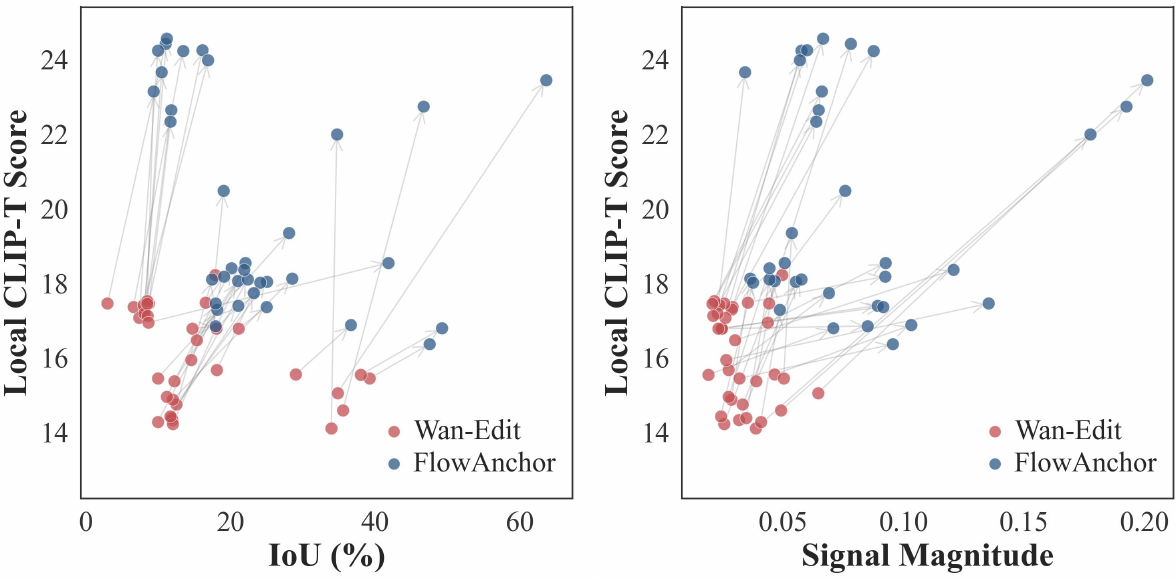}
    \vspace{-1.5 mm}
    \caption{
    \textbf{Editing Signal: Ours \vs Wan-Edit~\cite{li2025five}.}
    We compare the editing signal $\Delta V$ across diverse cases, with gray lines connecting paired results for the same instance.
    Our method exhibits higher IoU (left) and stronger magnitude (right), indicating precise localization and robust signal strength.
    %
    Consequently, these improvements result in higher Local CLIP-T scores, demonstrating superior editing performance.
    }
    \label{fig:scatter_plot}
    \vspace{-2 mm}
\end{figure}

\subsection{Quantitative Verification of Editing Signal Stability} To validate our solution to the issues in \subsecref{motivation}, we provide further comparisons against Wan-Edit~\cite{li2025five}.
%
%
%
%
Quantitative results shown in~\figref{scatter_plot} reveal two critical improvements:
\textbf{(1) Improved Localization:} Our method achieves higher IoU against ground-truth masks, concentrating the editing signal on the target region to induce precise semantic changes.
\textbf{(2) Enhanced Signal Magnitude:} Our method sustains significantly higher signal magnitude, ensuring sufficient strength to drive the editing trajectory.
Consequently, these enhancements translate to consistently higher Local CLIP-T scores, confirming the effectiveness of \textbf{FlowAnchor} in stabilizing the editing signal, ultimately yielding superior editing performance.

\subsection{Robustness to Mask Granularity}
\label{subsec:mask}
\begin{figure}[!t]
    \centering
    \includegraphics[width=0.95\columnwidth]{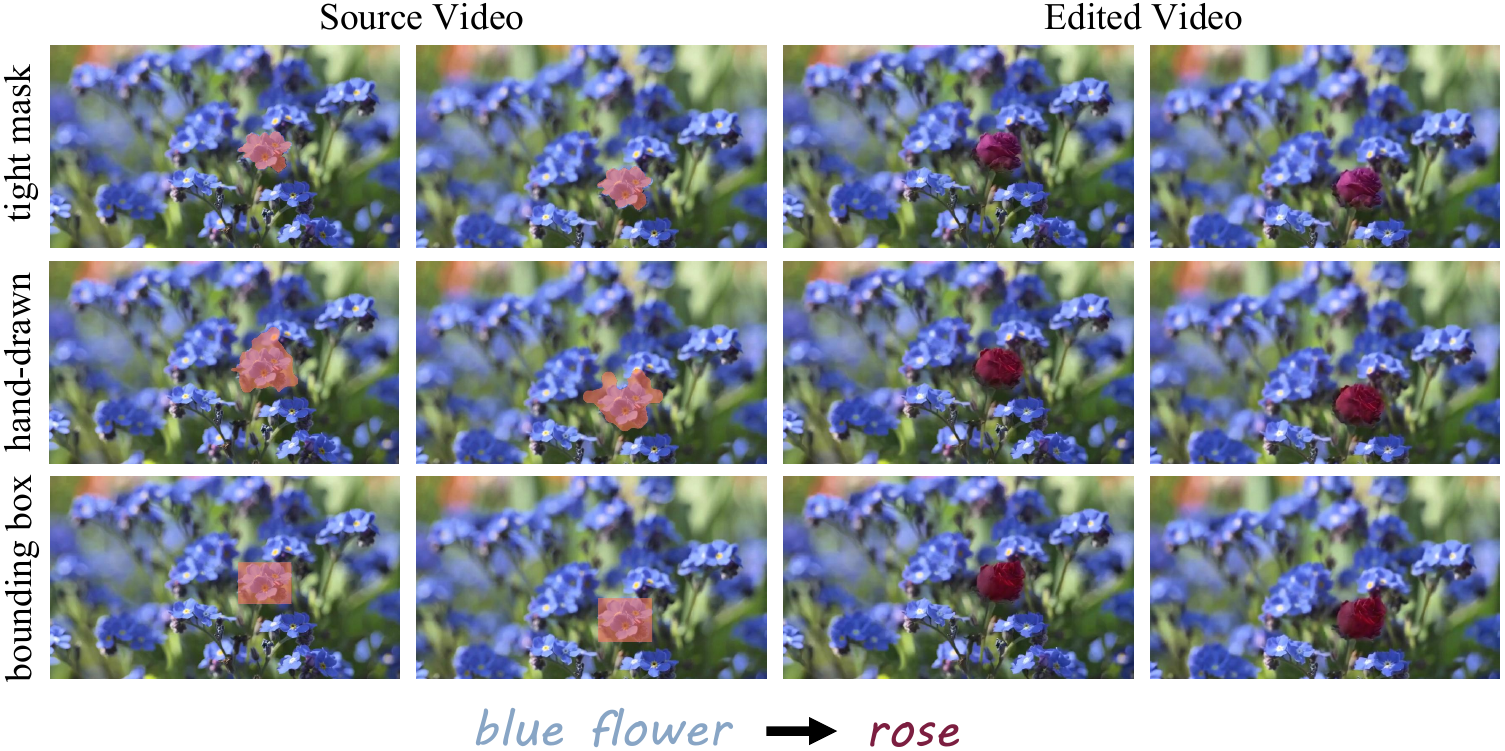}
    \vskip -0.1 in
    \caption{\textbf{Robustness to mask granularity.} FlowAnchor produces highly consistent edits across various mask granularities, ranging from tight masks to free-form hand-drawn scribbles and coarse bounding-boxes. The colored regions in the source video denote the masks. This suggests that FlowAnchor does not rely on pixel-accurate mask annotations, making it more practical for real-world interactive editing.}

    \label{fig:mask}
    \vskip -0.05in
\end{figure}

To demonstrate its robustness to mask precision, we evaluate FlowAnchor using masks of varying granularity, including tight segmentation, hand-drawn scribbles, and coarse bounding boxes.
%
As illustrated in \figref{mask}, our method maintains visually consistent editing quality across all conditions.
This inherent tolerance to imprecision stems from our design: the mask serves only as a spatial anchor during the early denoising steps and is decoupled from the later detail-generation stages.
Consequently, FlowAnchor eliminates the need for pixel-accurate guidance, making it highly practical for real-world interactive editing.

\subsection{Limitations and Future Work} 
\begin{figure}[!t]
    \centering
    \includegraphics[width=0.95\columnwidth]{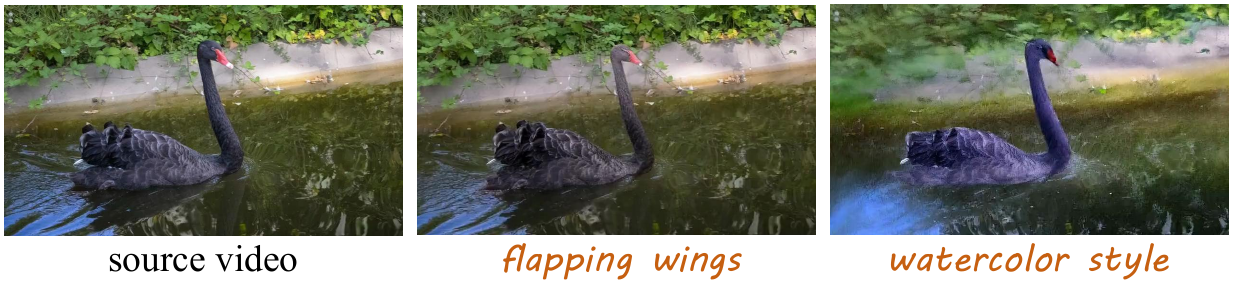}
    \vskip -0.05in
    \caption{\textbf{Limitations in global style transfer and motion editing.}}
    \label{fig:limitation}
    \vspace{-2 mm}
\end{figure}

Although our method shows strong performance across diverse editing tasks, it still struggles with global style transformations and substantial motion changes, which are inherited from the inversion-free paradigm~\cite{kulikov2025flowedit}, as shown in~\figref{limitation}.
We leave addressing these challenges as an important direction for future work.

\section{Conclusion}
\label{sec:conclusion}

In this work, we present \textbf{FlowAnchor}, a training-free framework that stabilizes the editing signal in inversion-free flow-based video editing.
We identify the challenge of \textit{instability in the editing signal}: imprecise localization and weakened magnitude, which leads to distorted editing trajectories and degraded editing results.
To address this, we introduce Spatial-aware Attention Refinement (SAR) and Adaptive Magnitude Modulation (AMM) to spatially anchor and adaptively strengthen the signal, which jointly enable a stable editing trajectory.
Qualitative and quantitative results demonstrate that FlowAnchor consistently outperforms existing methods across diverse editing scenarios, effectively advancing the capability of inversion-free video editing.

\bibliographystyle{ACM-Reference-Format}
\bibliography{main}

@String(ECCV= {Eur. Conf. Comput. Vis.})

@String(SIGGRAPH={SIGGRAPH})

@String(ECCV  = {ECCV})

@String{Computer = "{IEEE} Computer" }

@String{Springer = "Springer-Verlag" }

@inproceedings{geyertokenflow,
  title={TokenFlow: Consistent Diffusion Features for Consistent Video Editing},
  author={Geyer, Michal and Bar-Tal, Omer and Bagon, Shai and Dekel, Tali},
  booktitle={The Twelfth International Conference on Learning Representations},
  year={2024}
}

@inproceedings{li2025five,
  title={Five-bench: A fine-grained video editing benchmark for evaluating emerging diffusion and rectified flow models},
  author={Li, Minghan and Xie, Chenxi and Wu, Yichen and Zhang, Lei and Wang, Mengyu},
  booktitle={Proceedings of the IEEE/CVF International Conference on Computer Vision},
  pages={16672--16681},
  year={2025}
}

@inproceedings{wang2025taming,
  title={Taming Rectified Flow for Inversion and Editing},
  author={Wang, Jiangshan and Pu, Junfu and Qi, Zhongang and Guo, Jiayi and Ma, Yue and Huang, Nisha and Chen, Yuxin and Li, Xiu and Shan, Ying},
  booktitle={International Conference on Machine Learning},
  pages={64044--64058},
  year={2025},
  organization={PMLR}
}

@inproceedings{wang2025videodirector,
  title={Videodirector: Precise video editing via text-to-video models},
  author={Wang, Yukun and Wang, Longguang and Ma, Zhiyuan and Hu, Qibin and Xu, Kai and Guo, Yulan},
  booktitle={Proceedings of the IEEE/CVF Conference on Computer Vision and Pattern Recognition},
  pages={2589--2598},
  year={2025}
}

@inproceedings{kulikov2025flowedit,
  title={Flowedit: Inversion-free text-based editing using pre-trained flow models},
  author={Kulikov, Vladimir and Kleiner, Matan and Huberman-Spiegelglas, Inbar and Michaeli, Tomer},
  booktitle={Proceedings of the IEEE/CVF International Conference on Computer Vision},
  pages={19721--19730},
  year={2025}
}

@article{ho2022video,
  title={Video diffusion models},
  author={Ho, Jonathan and Salimans, Tim and Gritsenko, Alexey and Chan, William and Norouzi, Mohammad and Fleet, David J},
  journal={Advances in neural information processing systems},
  volume={35},
  pages={8633--8646},
  year={2022}
}

@inproceedings{singermake,
  title={Make-A-Video: Text-to-Video Generation without Text-Video Data},
  author={Singer, Uriel and Polyak, Adam and Hayes, Thomas and Yin, Xi and An, Jie and Zhang, Songyang and Hu, Qiyuan and Yang, Harry and Ashual, Oron and Gafni, Oran and others},
  booktitle={The Eleventh International Conference on Learning Representations},
  year={2023}
}

@inproceedings{rombach2022high,
  title={High-resolution image synthesis with latent diffusion models},
  author={Rombach, Robin and Blattmann, Andreas and Lorenz, Dominik and Esser, Patrick and Ommer, Bj{\"o}rn},
  booktitle={Proceedings of the IEEE/CVF conference on computer vision and pattern recognition},
  pages={10684--10695},
  year={2022}
}

@inproceedings{yangcogvideox,
  title={CogVideoX: Text-to-Video Diffusion Models with An Expert Transformer},
  author={Yang, Zhuoyi and Teng, Jiayan and Zheng, Wendi and Ding, Ming and Huang, Shiyu and Xu, Jiazheng and Yang, Yuanming and Hong, Wenyi and Zhang, Xiaohan and Feng, Guanyu and others},
  booktitle={The Thirteenth International Conference on Learning Representations},
  year={2025}
}

@article{kong2024hunyuanvideo,
  title={Hunyuanvideo: A systematic framework for large video generative models},
  author={Kong, Weijie and Tian, Qi and Zhang, Zijian and Min, Rox and Dai, Zuozhuo and Zhou, Jin and Xiong, Jiangfeng and Li, Xin and Wu, Bo and Zhang, Jianwei and others},
  journal={arXiv preprint arXiv:2412.03603},
  year={2024}
}

@article{wan2025wan,
  title={Wan: Open and advanced large-scale video generative models},
  author={Wan, Team and Wang, Ang and Ai, Baole and Wen, Bin and Mao, Chaojie and Xie, Chen-Wei and Chen, Di and Yu, Feiwu and Zhao, Haiming and Yang, Jianxiao and others},
  journal={arXiv preprint arXiv:2503.20314},
  year={2025}
}

@inproceedings{peebles2023scalable,
  title={Scalable diffusion models with transformers},
  author={Peebles, William and Xie, Saining},
  booktitle={Proceedings of the IEEE/CVF international conference on computer vision},
  pages={4195--4205},
  year={2023}
}

@inproceedings{qi2023fatezero,
  title={Fatezero: Fusing attentions for zero-shot text-based video editing},
  author={Qi, Chenyang and Cun, Xiaodong and Zhang, Yong and Lei, Chenyang and Wang, Xintao and Shan, Ying and Chen, Qifeng},
  booktitle={Proceedings of the IEEE/CVF International Conference on Computer Vision},
  pages={15932--15942},
  year={2023}
}

@inproceedings{ceylan2023pix2video,
  title={Pix2video: Video editing using image diffusion},
  author={Ceylan, Duygu and Huang, Chun-Hao P and Mitra, Niloy J},
  booktitle={Proceedings of the IEEE/CVF international conference on computer vision},
  pages={23206--23217},
  year={2023}
}

@inproceedings{yang2023rerender,
  title={Rerender a video: Zero-shot text-guided video-to-video translation},
  author={Yang, Shuai and Zhou, Yifan and Liu, Ziwei and Loy, Chen Change},
  booktitle={SIGGRAPH Asia 2023 Conference Papers},
  pages={1--11},
  year={2023}
}

@inproceedings{congflatten,
  title={FLATTEN: optical FLow-guided ATTENtion for consistent text-to-video editing},
  author={Cong, Yuren and Xu, Mengmeng and Chen, Shoufa and Ren, Jiawei and Xie, Yanping and Perez-Rua, Juan-Manuel and Rosenhahn, Bodo and Xiang, Tao and He, Sen and others},
  booktitle={The Twelfth International Conference on Learning Representations},
  year={2024}
}

@article{zhang2023controlvideo,
  title={ControlVideo: Training-free Controllable Text-to-Video Generation},
  author={Zhang, Yabo and Wei, Yuxiang and Jiang, Dongsheng and Zhang, Xiaopeng and Zuo, Wangmeng and Tian, Qi},
  journal={arXiv preprint arXiv:2305.13077},
  year={2023}
}

@inproceedings{kara2024rave,
  title={Rave: Randomized noise shuffling for fast and consistent video editing with diffusion models},
  author={Kara, Ozgur and Kurtkaya, Bariscan and Yesiltepe, Hidir and Rehg, James M and Yanardag, Pinar},
  booktitle={Proceedings of the IEEE/CVF Conference on Computer Vision and Pattern Recognition},
  pages={6507--6516},
  year={2024}
}

@inproceedings{yang2025videograin,
  title={Videograin: Modulating space-time attention for multi-grained video editing},
  author={Yang, Xiangpeng and Zhu, Linchao and Fan, Hehe and Yang, Yi},
  booktitle={The Thirteenth International Conference on Learning Representations},
  year={2025}
}

@article{jiao2025uniedit,
  title={Uniedit-flow: Unleashing inversion and editing in the era of flow models},
  author={Jiao, Guanlong and Huang, Biqing and Wang, Kuan-Chieh and Liao, Renjie},
  journal={arXiv preprint arXiv:2504.13109},
  year={2025}
}

@inproceedings{lipmanflow,
  title={Flow Matching for Generative Modeling},
  author={Lipman, Yaron and Chen, Ricky TQ and Ben-Hamu, Heli and Nickel, Maximilian and Le, Matthew},
  booktitle={The Eleventh International Conference on Learning Representations},
  year={2023}
}

@inproceedings{liuflow,
  title={Flow Straight and Fast: Learning to Generate and Transfer Data with Rectified Flow},
  author={Liu, Xingchao and Gong, Chengyue and others},
  booktitle={The Eleventh International Conference on Learning Representations},
  year={2023}
}

@inproceedings{xu2024inversion,
  title={Inversion-free image editing with language-guided diffusion models},
  author={Xu, Sihan and Huang, Yidong and Pan, Jiayi and Ma, Ziqiao and Chai, Joyce},
  booktitle={Proceedings of the IEEE/CVF Conference on Computer Vision and Pattern Recognition},
  pages={9452--9461},
  year={2024}
}

@inproceedings{yoonsplitflow,
  title={SplitFlow: Flow Decomposition for Inversion-Free Text-to-Image Editing},
  author={Yoon, Sung-Hoon and Li, Minghan and Beaudouin, Gaspard and Wen, Congcong and Azhar, Muhammad Rafay and Wang, Mengyu},
  booktitle={The Thirty-ninth Annual Conference on Neural Information Processing Systems},
  year={2025}
}

@article{kim2025flowalign,
  title={Flowalign: Trajectory-regularized, inversion-free flow-based image editing},
  author={Kim, Jeongsol and Hong, Yeobin and Park, Jonghyun and Ye, Jong Chul},
  journal={arXiv preprint arXiv:2505.23145},
  year={2025}
}

@article{cai2025dfvedit,
  title={DFVEdit: Conditional Delta Flow Vector for Zero-shot Video Editing},
  author={Cai, Lingling and Zhao, Kang and Yuan, Hangjie and Wang, Xiang and Zhang, Yingya and Huang, Kejie},
  journal={arXiv preprint arXiv:2506.20967},
  year={2025}
}

@article{li2025flowdirector,
  title={Flowdirector: Training-free flow steering for precise text-to-video editing},
  author={Li, Guangzhao and Yang, Yanming and Song, Chenxi and Zhang, Chi},
  journal={arXiv preprint arXiv:2506.05046},
  year={2025}
}

@article{kong2025taming,
  title={Taming flow-based i2v models for creative video editing},
  author={Kong, Xianghao and Chen, Hansheng and Guo, Yuwei and Zhang, Lvmin and Wetzstein, Gordon and Agrawala, Maneesh and Rao, Anyi},
  journal={arXiv preprint arXiv:2509.21917},
  year={2025}
}

@inproceedings{perazzi2016benchmark,
  title={A benchmark dataset and evaluation methodology for video object segmentation},
  author={Perazzi, Federico and Pont-Tuset, Jordi and McWilliams, Brian and Van Gool, Luc and Gross, Markus and Sorkine-Hornung, Alexander},
  booktitle={Proceedings of the IEEE conference on computer vision and pattern recognition},
  pages={724--732},
  year={2016}
}

@inproceedings{kirillov2023segment,
  title={Segment anything},
  author={Kirillov, Alexander and Mintun, Eric and Ravi, Nikhila and Mao, Hanzi and Rolland, Chloe and Gustafson, Laura and Xiao, Tete and Whitehead, Spencer and Berg, Alexander C and Lo, Wan-Yen and others},
  booktitle={Proceedings of the IEEE/CVF international conference on computer vision},
  pages={4015--4026},
  year={2023}
}

@inproceedings{radford2021learning,
  title={Learning transferable visual models from natural language supervision},
  author={Radford, Alec and Kim, Jong Wook and Hallacy, Chris and Ramesh, Aditya and Goh, Gabriel and Agarwal, Sandhini and Sastry, Girish and Askell, Amanda and Mishkin, Pamela and Clark, Jack and others},
  booktitle={International conference on machine learning},
  pages={8748--8763},
  year={2021},
  organization={PmLR}
}

@article{oquab2024dinov2,
  title={DINOv2: Learning Robust Visual Features without Supervision},
  author={Oquab, Maxime and Darcet, Timoth{\'e}e and Moutakanni, Th{\'e}o and Vo, Huy and Szafraniec, Marc and Khalidov, Vasil and Fernandez, Pierre and Haziza, Daniel and Massa, Francisco and El-Nouby, Alaaeldin and others},
  journal={Transactions on Machine Learning Research Journal},
  year={2024}
}

@article{huynh2008scope,
  title={Scope of validity of PSNR in image/video quality assessment},
  author={Huynh-Thu, Quan and Ghanbari, Mohammed},
  journal={Electronics letters},
  volume={44},
  number={13},
  pages={800--801},
  year={2008},
  publisher={IET}
}

@inproceedings{lai2018learning,
  title={Learning blind video temporal consistency},
  author={Lai, Wei-Sheng and Huang, Jia-Bin and Wang, Oliver and Shechtman, Eli and Yumer, Ersin and Yang, Ming-Hsuan},
  booktitle={Proceedings of the European conference on computer vision (ECCV)},
  pages={170--185},
  year={2018}
}

@inproceedings{jiang2025vace,
  title={Vace: All-in-one video creation and editing},
  author={Jiang, Zeyinzi and Han, Zhen and Mao, Chaojie and Zhang, Jingfeng and Pan, Yulin and Liu, Yu},
  booktitle={Proceedings of the IEEE/CVF International Conference on Computer Vision},
  pages={17191--17202},
  year={2025}
}

@inproceedings{teed2020raft,
  title={Raft: Recurrent all-pairs field transforms for optical flow},
  author={Teed, Zachary and Deng, Jia},
  booktitle={European conference on computer vision},
  pages={402--419},
  year={2020},
  organization={Springer}
}

\clearpage
\appendix
\section{Summary}

In this supplementary material, we provide additional technical details, benchmark descriptions, and qualitative analyses.
The contents are organized as follows:
\begin{itemize}
    \item In \cref{sec:details}, we present the full implementation details of FlowAnchor, including the editing algorithm, the hyperparameter settings, and the concrete formulations of SAR and AMM.
    
    \item In \cref{sec:anchorbench}, we describe Anchor-Bench in detail, including the data collection pipeline, prompt and mask annotation process, and the definitions of all evaluation metrics.
    
    \item In \cref{sec:baseline_details}, we provide the reproduction details of all compared baseline methods and clarify the method-specific adaptations used for fair comparison.
    
    \item In \cref{sec:ablation}, we report additional ablation results, including the sensitivity of SAR and AMM to their hyperparameters and the effect of the SAR application range.
    
    \item In \cref{subsec:mask}, we evaluate the robustness of FlowAnchor to different mask granularities and show that the method remains effective even with coarse user inputs.
    
    \item In \cref{sec:addition}, we provide additional qualitative comparisons with FlowDirector~\cite{li2025flowdirector} and discuss its limitations in spatial localization and temporal stability.
    
    \item In \cref{sec:inpainting}, we further compare FlowAnchor with the representative inpainting-based method VACE~\cite{jiang2025vace}, highlighting the differences in editing completeness, texture preservation, object replacement, and robustness to deformation.
    
    \item In \cref{sec:more_results}, we present more qualitative results of FlowAnchor on diverse localized video editing scenarios, including texture editing, object replacement, object addition, and non-rigid transformation.
\end{itemize}

\section{Implementation Details of FlowAnchor}
\label{sec:details}

We build FlowAnchor on Wan2.1-T2V-1.3B~\cite{wan2025wan} and follow FlowEdit~\cite{kulikov2025flowedit} for the rectified-flow sampling formulation.
As in FlowEdit, we inherit the two sampling hyperparameters $n_{\max}$ and $n_{\text{avg}}$.
In all experiments, we set $T=25$ and use $n_{\max}=23$, i.e., the first two denoising steps are skipped and editing is performed over the remaining $23$ steps.
Skipping the earliest iterations helps preserve the coarse spatial structure of the source video, while still leaving sufficient room for the editing signal to steer the trajectory.
We set $n_{\text{avg}}=1$ and compute the editing signal once at each step for efficiency.

\begin{algorithm}[tb]
\caption{FlowAnchor Editing}
\label{alg:flowanchor}
\KwIn{Source video $X^{\mathrm{src}}$, source prompt $\mathcal{P}$, target prompt $\mathcal{P}^{*}$, mask $M$, time grid $\{t_i\}_{i=0}^{T}$, strengths $\beta_1,\beta_2,\gamma$, reference latent length $F_0$}
\KwOut{Edited video $Z^{\mathrm{edit}}_{0}$}

\BlankLine
$Z^{\mathrm{edit}}_{t_T} \leftarrow X^{\mathrm{src}}$\;
$\tau \leftarrow 0.6T$, \quad $\gamma_F \leftarrow \gamma \cdot \log(F)/\log(F_0)$\;

\For{$i=T,\dots,1$}{
    $N_{t_i} \sim \mathcal{N}(0,I)$\;
    $Z^{\mathrm{src}}_{t_i} \leftarrow (1-t_i)X^{\mathrm{src}} + t_iN_{t_i}$\;
    $Z^{\mathrm{tar}}_{t_i} \leftarrow Z^{\mathrm{edit}}_{t_i} + Z^{\mathrm{src}}_{t_i} - X^{\mathrm{src}}$\;

    \tcbox[
        colback=sarback,
        colframe=sarframe,
        colbacktitle=sarframe!55!white,
        coltitle=black,
        title=Spatial-aware Attention Refinement (SAR),
        fonttitle=\bfseries\small,
        boxrule=0.4pt,
        arc=2.5pt,
        left=2pt,right=2pt,top=1pt,bottom=1pt
    ]{
        \begin{minipage}{0.84\linewidth}
        \uIf{$t_i \ge \tau$}{
            $V^{\mathrm{tar}}_{t_i} \leftarrow V_{\mathrm{SAR}}(Z^{\mathrm{tar}}_{t_i}, t_i, \mathcal{P}^{*}, M, J_{\mathrm{tar}}, \beta_1, \beta_2)$\;
        }
        \Else{
            $V^{\mathrm{tar}}_{t_i} \leftarrow V(Z^{\mathrm{tar}}_{t_i}, t_i, \mathcal{P}^{*})$\;
        }
        \end{minipage}
    }

    $V^{\mathrm{src}}_{t_i} \leftarrow V(Z^{\mathrm{src}}_{t_i}, t_i, \mathcal{P})$\;
    $\Delta V_{t_i} \leftarrow V^{\mathrm{tar}}_{t_i} - V^{\mathrm{src}}_{t_i}$\;

    \tcbox[
        colback=ammback,
        colframe=ammframe,
        colbacktitle=ammframe!55!white,
        coltitle=black,
        title=Adaptive Magnitude Modulation (AMM),
        fonttitle=\bfseries\small,
        boxrule=0.4pt,
        arc=2.5pt,
        left=2pt,right=2pt,top=1pt,bottom=1pt
    ]{
        \begin{minipage}{0.84\linewidth}
        $C_{t_i} \leftarrow \mathrm{Norm}(\Delta V_{t_i})$\;
        $\Delta V_{t_i}^{\mathrm{AMM}} \leftarrow (1+\gamma_F \cdot C_{t_i}) \odot \Delta V_{t_i}$\;
        \end{minipage}
    }

    $Z^{\mathrm{edit}}_{t_{i-1}} \leftarrow Z^{\mathrm{edit}}_{t_i} + (t_{i-1}-t_i)\Delta V_{t_i}^{\mathrm{AMM}}$\;
}
\Return{$Z^{\mathrm{edit}}_{0}$}
\end{algorithm}

\subsection{SAR Implementation}

SAR is applied to all $30$ cross-attention (CA) layers during the early denoising stage, i.e., $t \in [T,\tau]$ with $\tau=0.6T$.
Unless otherwise specified, we fix the two modulation strengths to $\beta_1=\beta_2=0.3$.
SAR is applied to CA logits before the softmax operation.
Concretely, let $A^{(l)}\in\mathbb{R}^{N_l\times L}$ denote the CA logits at layer $l$, where $N_l=F_lH_lW_l$ is the number of spatio-temporal latent tokens and $L$ is the number of text tokens.
The normalized CA weights are obtained by applying softmax along the text-token dimension, i.e., for each spatio-temporal token $i$,
\begin{equation}
\tilde{A}^{(l)}_{i,j}
=
\frac{\exp(A^{(l)}_{i,j})}{\sum_{k=1}^{L}\exp(A^{(l)}_{i,k})},
\end{equation}
such that the attention weights over all text tokens sum to one for each fixed $i$.
Given the target token set $J_{\mathrm{tar}}$ and the binary spatial anchor mask $M\in\{0,1\}^{N_l}$ at the corresponding latent resolution, SAR first performs text-token modulation inside the masked region:
\begin{equation}
A'_{i,j}=
\begin{cases}
A_{i,j}+\beta_1(A_i^{\max}-A_{i,j}), & M_i=1,\ j\in J_{\mathrm{tar}},\\
A_{i,j}-\beta_1(A_{i,j}-A_i^{\min}), & M_i=1,\ j\notin J_{\mathrm{tar}},\\
A_{i,j}, & \text{otherwise},
\end{cases}
\label{eq:sar_step1_appendix}
\end{equation}
where
\begin{equation}
A_i^{\max}=\max_k A_{i,k}, \qquad
A_i^{\min}=\min_k A_{i,k}.
\end{equation}
This step increases the relative dominance of the target tokens within the masked region while suppressing interference from irrelevant text tokens.

A second modulation is then applied along the spatio-temporal dimension for target tokens:
\begin{equation}
A''_{i,j}=
\begin{cases}
A'_{i,j}+\beta_2(A_j^{\prime\max}-A'_{i,j}), & M_i=1,\ j\in J_{\mathrm{tar}},\\
A'_{i,j}-\beta_2(A'_{i,j}-A_j^{\prime\min}), & M_i=0,\ j\in J_{\mathrm{tar}},\\
A'_{i,j}, & \text{otherwise},
\end{cases}
\label{eq:sar_step2_appendix}
\end{equation}
where
\begin{equation}
A_j^{\prime\max}=\max_p A'_{p,j}, \qquad
A_j^{\prime\min}=\min_p A'_{p,j}.
\end{equation}
The refined logits $A''$ are then passed to softmax to obtain the final normalized CA map.

The above formulation preserves numerical stability.
For $\beta_1,\beta_2\in[0,1]$, each update is a convex interpolation toward an existing maximum or minimum value.
For example, when $M_i=1$ and $j\in J_{\mathrm{tar}}$,
\begin{equation}
A'_{i,j}=(1-\beta_1)A_{i,j}+\beta_1A_i^{\max},
\end{equation}
and when $M_i=1$ and $j\notin J_{\mathrm{tar}}$,
\begin{equation}
A'_{i,j}=(1-\beta_1)A_{i,j}+\beta_1A_i^{\min}.
\end{equation}
Therefore,
\begin{equation}
A_i^{\min}\leq A'_{i,j}\leq A_i^{\max}.
\end{equation}
By the same argument, the second-step modulation also satisfies
\begin{equation}
A_j^{\prime\min}\leq A''_{i,j}\leq A_j^{\prime\max}.
\end{equation}
Hence SAR does not introduce values outside the original logit range, but only reshapes their relative contrast.
Since the normalization is still performed by softmax after modulation, the resulting attention remains a valid probability distribution.

\begin{figure*}[t]
    \centering
    \includegraphics[width=\textwidth]{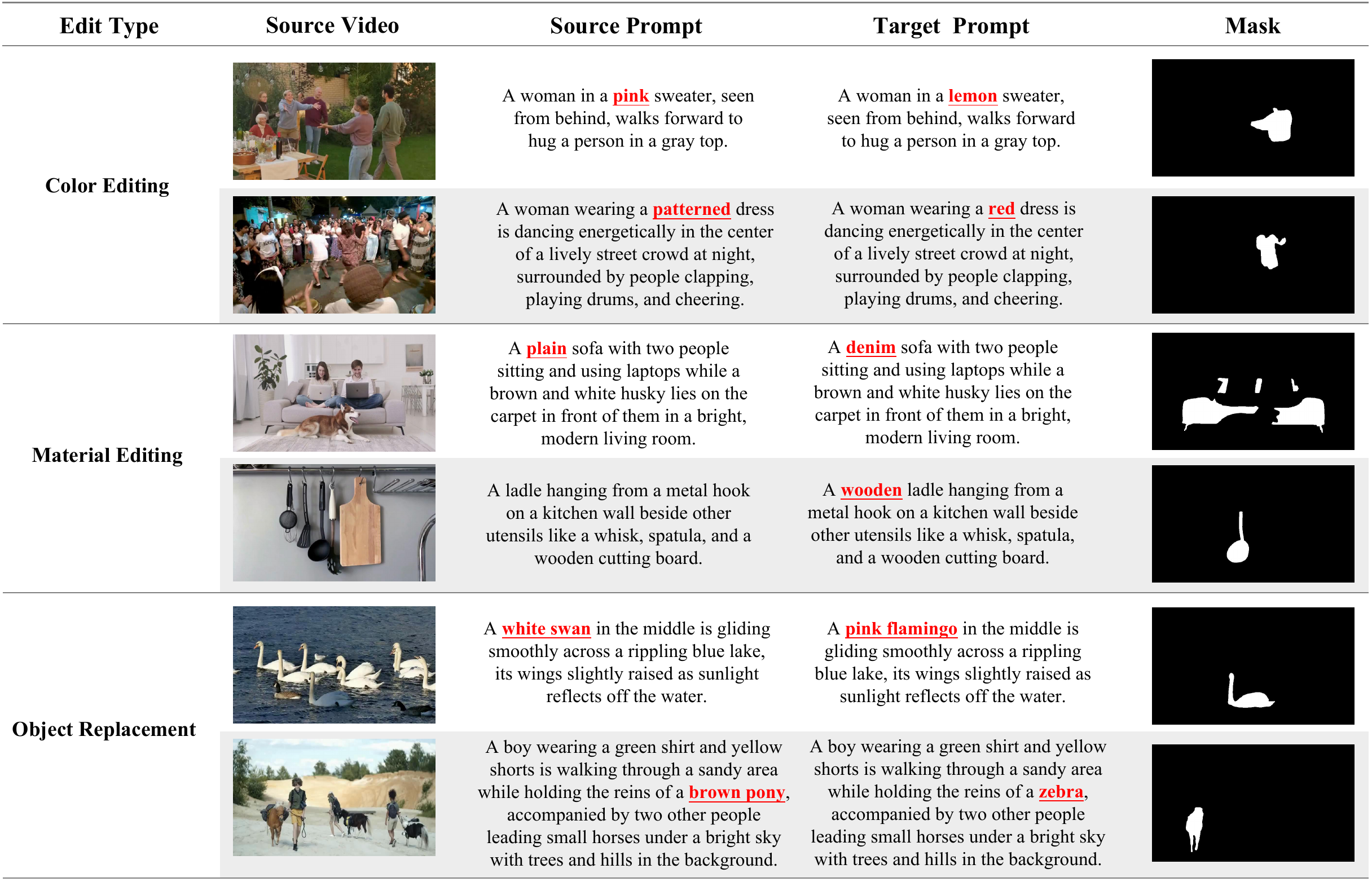}
    \vskip -0.1in
    \caption{\textbf{Examples and annotations in Anchor-Bench.}
    Anchor-Bench covers three localized editing types, including \textbf{color editing}, \textbf{material editing}, and \textbf{object replacement}.
    The edited tokens are highlighted in red to indicate the semantic modification.
    The masks specify the target editing regions for localized evaluation.}
    \label{fig:anchorbench_examples}
\end{figure*}

\begin{figure*}[t]
    \centering
    \includegraphics[width=\textwidth]{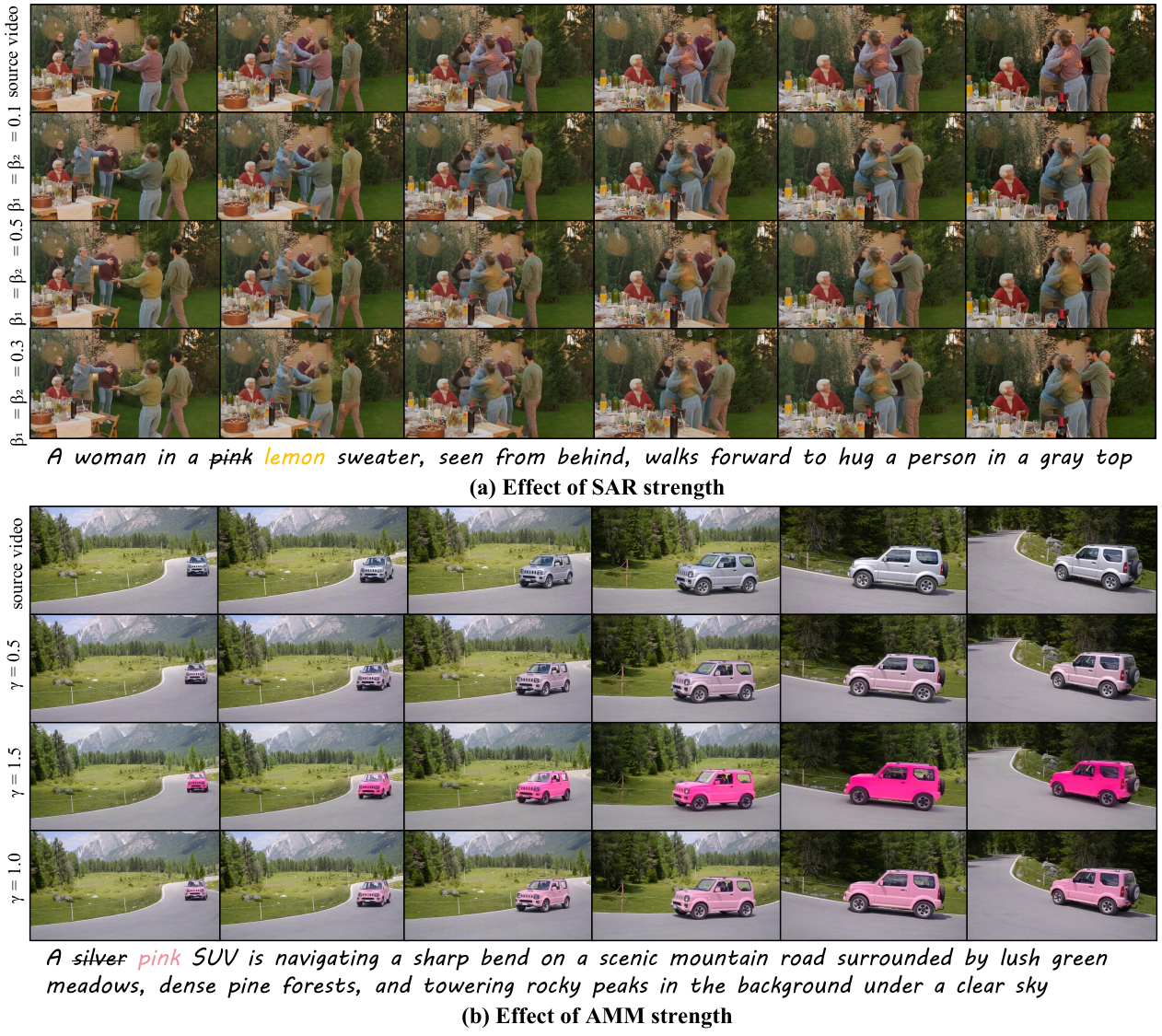}
    \vskip -0.1in
    \caption{\textbf{Ablation on hyperparameters of SAR and AMM.}
    (a) Effect of SAR strengths \((\beta_1,\beta_2)\).
    Smaller values lead to insufficient attention modulation, while larger values may introduce instability.
    \(\beta_1=\beta_2=0.3\) achieves a good balance.
    (b) Effect of AMM strength \(\gamma\).
    Smaller \(\gamma\) leads to under-editing, while larger \(\gamma\) causes over-editing and structural distortion.
    \(\gamma=1.0\) provides the best trade-off between editing strength and structural fidelity.}
    \label{fig:ablation_sar_amm}
\end{figure*}

\subsection{AMM Implementation}

AMM is applied at every denoising step.
Let the editing signal at timestep $t_i$ be
\begin{equation}
\Delta V_{t_i}\in\mathbb{R}^{B\times C\times F\times H\times W},
\end{equation}
where $B$ is the batch size, $C$ is the channel dimension, and $F,H,W$ are the latent temporal and spatial resolutions.
To obtain the contrast map $C_{t_i}$ used in AMM, we first average the editing signal over the channel dimension:
\begin{equation}
\bar{V}_{t_i}=\frac{1}{C}\sum_{c=1}^{C}\Delta V_{t_i}^{(c)}
\in\mathbb{R}^{B\times 1\times F\times H\times W}.
\end{equation}
We then perform min-max normalization \emph{independently for each sample} over all spatio-temporal positions:
\begin{equation}
C_{t_i}^{(b)}=
\frac{
\bar{V}_{t_i}^{(b)}-\min\!\left(\bar{V}_{t_i}^{(b)}\right)
}{
\max\!\left(\bar{V}_{t_i}^{(b)}\right)-\min\!\left(\bar{V}_{t_i}^{(b)}\right)+\epsilon
},
\qquad b=1,\dots,B,
\label{eq:cti_appendix}
\end{equation}
where the $\min$ and $\max$ are computed over the flattened $F\times H\times W$ dimension and $\epsilon=10^{-7}$ is used for numerical stability.
Thus, $C_{t_i}\in[0,1]^{B\times 1\times F\times H\times W}$ is a sample-wise normalized dynamic mask, which is broadcast along the channel dimension when modulating the editing signal.

The frame-adaptive amplification factor is defined as
\begin{equation}
\gamma_F=\gamma\cdot\frac{\log F}{\log F_0}.
\end{equation}
The final modulated editing signal is
\begin{equation}
\Delta \tilde{V}_{t_i}
=
\left(1+\gamma_F\, C_{t_i}\right)\odot \Delta V_{t_i},
\label{eq:amm_appendix}
\end{equation}
where $\odot$ denotes element-wise multiplication with broadcasting over the channel dimension.
we set $\gamma=1.0$.

The choice of $F_0=21$ follows the native temporal scale of Wan2.1~\cite{wan2025wan} in latent space.
Wan2.1 uses $81$ frames as the default maximum video length in pixel space, and its VAE applies $4\times$ temporal downsampling.
Therefore, the corresponding latent temporal length is
\begin{equation}
F_0=\frac{81-1}{4}+1=21.
\end{equation}
Using $F_0=21$ makes the amplification factor consistent with the default temporal resolution at which the editing signal is actually computed.

Eq.~\eqref{eq:cti_appendix} also makes the modulation numerically stable.
Since $C_{t_i}\in[0,1]$, the amplification coefficient in Eq.~\eqref{eq:amm_appendix} is bounded by
\begin{equation}
1 \le 1+\gamma_F C_{t_i}\le 1+\gamma_F.
\end{equation}
Therefore, AMM only rescales the editing signal within a controlled range and does not cause unbounded amplification.

\section{Anchor-Bench}
\label{sec:anchorbench}

\subsection{Dataset}

FiVE-Bench~\cite{li2025five} provides a valuable benchmark for fine-grained video editing with object-level prompts and masks.
However, it is less focused on challenging localized edits in multi-object videos.
To better evaluate localized video editing in more realistic scenarios, we construct \textbf{Anchor-Bench}, a benchmark consisting of $74$ text-video editing pairs.
All videos are collected from the Internet and cover diverse real-world scenes with multiple objects, cluttered backgrounds, and fast motion.
The benchmark contains videos of up to $81$ frames at $480$p resolution.

Anchor-Bench focuses on three localized editing categories:
\textit{(1) color editing},
\textit{(2) material editing}, and
\textit{(3) object replacement}, where object replacement includes both rigid and non-rigid objects.
For each source video, we annotate one source prompt and multiple target prompts corresponding to different local editing instructions.
We first use GPT-5 to generate candidate prompts and then manually refine them to ensure semantic correctness and unambiguous reference to the intended editing target.
In particular, when multiple similar objects or persons appear in the same scene, we explicitly add discriminative cues such as object category, color, relative position, or surrounding context, so that the edited subject can be uniquely identified from the prompt itself.
The source and target prompts are otherwise kept as consistent as possible, differing only in the edited attribute or object.

For each target prompt, we additionally provide a corresponding edit mask sequence for localized evaluation.
We manually annotate the target region on the first frame and propagate it to the remaining frames using optical flow~\cite{teed2020raft}.
Representative examples are shown in Fig.~\ref{fig:anchorbench_examples}.

\subsection{Evaluation Metrics}

For text--video alignment, we report both the global CLIP-T and the localized L.CLIP-T~\cite{radford2021learning}.
We use CLIP ViT-L/14 to compute all CLIP-based scores.
CLIP-T measures the global alignment between the edited video and the full target prompt.
L.CLIP-T focuses on the edited region by evaluating the cropped masked region against a local target phrase containing only the edited semantics.

For fidelity, we report both structure-level and pixel-level metrics.
At the structure level, we compute Local DINO similarity (L.DINO) using DINOv2 ViT-B/14~\cite{oquab2024dinov2}.
The cosine similarity is measured between the cropped source region and the cropped edited region, reflecting whether the local structure is preserved after editing.
At the pixel level, we report masked PSNR (M.PSNR)~\cite{huynh2008scope}, which evaluates reconstruction quality in the unedited regions.

For temporal consistency, we report CLIP-F~\cite{radford2021learning} and Warp-Err~\cite{lai2018learning}.
CLIP-F measures semantic continuity between consecutive frames using CLIP features.
Warp-Err measures pixel-level temporal stability by first estimating optical flow with RAFT~\cite{teed2020raft}, warping each edited frame to the next frame, and then computing the deviation between the warped frame and the generated frame.
Lower Warp-Err indicates better temporal consistency.

Different from FiVE-Bench, which computes metrics on sparsely sampled frames, we evaluate all frame-wise metrics on the full video sequence.
Although this protocol is more computationally expensive, it provides a more faithful assessment of local editing quality and temporal consistency throughout the video.

\begin{figure}[t]
    \centering
    \includegraphics[width=\columnwidth]{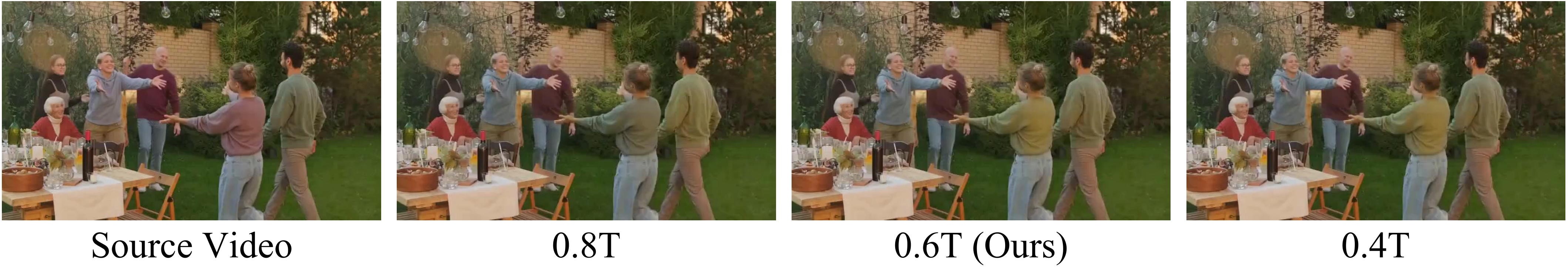}
    \vskip -0.05in
    \caption{\textbf{Effect of SAR application range.}
    We vary the cutoff timestep \(\tau\) for applying SAR.
    A short range (\(\tau=0.8T\)) provides insufficient semantic guidance, while \(\tau=0.6T\) yields the best localization quality.
    Further extending the range to \(\tau=0.4T\) brings no clear additional benefit.}
    \label{fig:timestep}
    \vskip -0.05in
    \vspace{-3mm}
\end{figure}
\begin{figure*}[t]
    \centering
    \includegraphics[width=\textwidth]{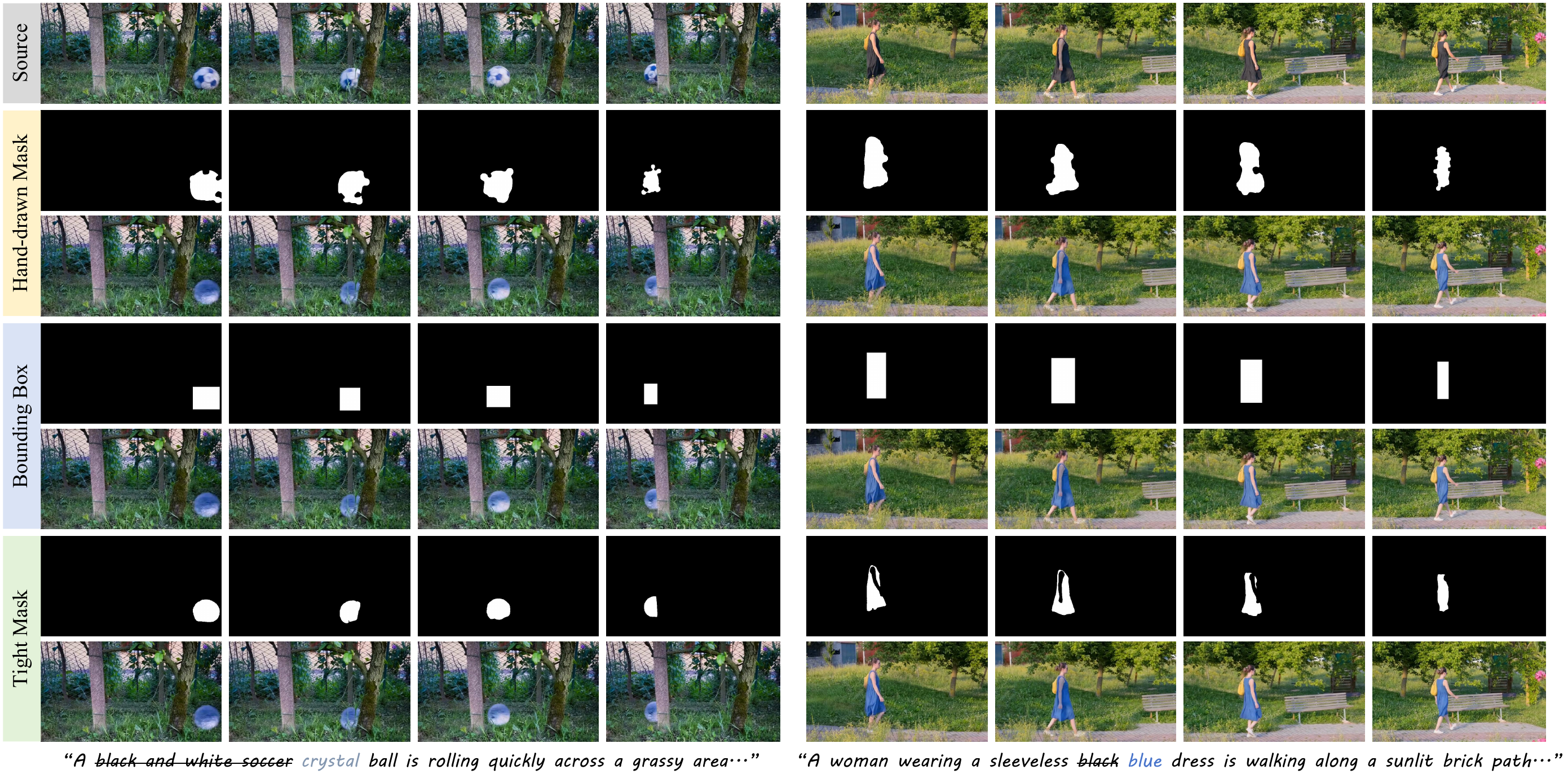}
    \vskip -0.1in
    \caption{\textbf{Comparison with the inpainting-based method VACE~\cite{jiang2025vace}.} 
    While VACE is a unified \emph{training-based} framework for masked video editing, it often exhibit \emph{under-editing} issues, failing to editing the whole masked region or even produce negligible effects.
    In contrast, our \emph{training-free} FlowAnchor accommodates diverse editing types, delivering highly precise edits that strictly align with the target text while maintaining structural stability.}
    \label{fig:mask_blue}
\end{figure*}

\section{Baseline Implementation Details}
\label{sec:baseline_details}

We compare FlowAnchor with TokenFlow~\cite{geyertokenflow}, 
VideoGrain~\cite{yang2025videograin}, 
RF-Solver-Edit~\cite{wang2025taming}, 
UniEdit-Flow~\cite{jiao2025uniedit}, 
Wan-Edit~\cite{li2025five}, 
and FlowDirector~\cite{li2025flowdirector}.
For all baselines, we use the official implementations and follow their default hyperparameter settings without additional tuning.

Among the compared methods, several baselines do not natively support explicit spatial grounding from benchmark masks.
To provide a stronger mask-guided baseline on top of FlowEdit-style video editing, we further construct \textbf{Wan-Edit+Mask} based on Wan-Edit~\cite{li2025five}.
Specifically, after each editing update, we perform latent blending between the edited latent and the source latent:
\begin{equation}
Z^{\mathrm{blend}}_{t_{i-1}}
=
M_{t_{i-1}} \odot Z^{\mathrm{edit}}_{t_{i-1}}
+
(1-M_{t_{i-1}}) \odot Z^{\mathrm{src}}_{t_{i-1}},
\end{equation}
where $M_{t_{i-1}}$ denotes the benchmark mask resized to the latent resolution at timestep $t_{i-1}$.

FlowDirector~\cite{li2025flowdirector} performs spatial control by extracting CA maps as implicit masks to gate the editing process.
However, we observe that such attention-based localization is often inaccurate in complex video scenarios, especially in multi-object scenes or under fast motion.
This behavior is consistent with our observations on Wan-Edit, where the CA maps can be spatially ambiguous and temporally unstable, leading to imprecise or drifting editing regions.

Finally, we emphasize that the improvements of FlowAnchor are not solely attributed to the use of explicit masks.
Instead, SAR and AMM directly operate on the model’s internal representations to enhance both the localization and the strength of the editing signal, enabling robust and consistent editing beyond mask guidance alone.

For fair comparison, all methods, including FlowAnchor and all mask-aware baselines, use the same benchmark masks during evaluation.
All quantitative results are computed under the same evaluation protocol and hardware setting.

\section{Additional Ablation Study}
\label{sec:ablation}

\subsection{Hyperparameter Analysis of SAR and AMM}

We analyze the sensitivity of FlowAnchor to the strengths of SAR and AMM.
As shown in \cref{fig:ablation_sar_amm}(a), smaller values of $\beta_1,\beta_2$ lead to insufficient attention modulation, resulting in weak localization of the target semantics.
Increasing the strengths improves semantic focus and makes the target region easier to localize.
However, further increasing $\beta_1,\beta_2$ beyond the default setting brings only marginal gains in localization, while making the editing more aggressive and slightly less favorable to overall fidelity in some cases.
Therefore, we choose $\beta_1=\beta_2=0.3$ as a robust default that achieves a good balance between effective localization and stable editing behavior.

As shown in \cref{fig:ablation_sar_amm}(b), the parameter $\gamma$ controls the magnitude of the editing signal.
A small $\gamma$ leads to under-editing, where the target semantics are not sufficiently expressed.
In contrast, a large $\gamma$ causes over-editing and structural distortion.
Empirically, $\gamma=1.0$ achieves the best trade-off between editing strength and structural fidelity.

\subsection{Effect of SAR Application Range}
We further study the timestep range where SAR is applied.
Recall that SAR is activated during the early denoising stage $t\in[T,\tau]$ to establish stable semantic localization.
As shown in \cref{fig:timestep}, a shorter application range with $\tau=0.8T$ provides insufficient guidance, leading to weaker localization of the target semantics.
Extending SAR to $\tau=0.6T$ significantly improves the editing effect and yields the most reliable results.
Further extending the application range to $\tau=0.4T$ does not bring clear additional gains, and may instead interfere with the later-stage generation of fine details.
This observation is consistent with the role of SAR in our framework: it mainly serves to anchor the editing region in the early stage, while the subsequent denoising steps are better left to preserve appearance and structural details.
Therefore, we adopt $\tau=0.6T$ as the default setting.

\begin{table}[!t]\centering
\caption{\textbf{Robustness to mask granularity.} 
We compare FlowAnchor using hand-drawn masks, coarse bounding boxes, and tight masks. 
Warp-Err is reported in $10^{-3}$. 
All metrics are evaluated using the tight mask protocol for fair comparison.}
\label{tab:mask_granularity}
\vspace{-2mm}
\renewcommand{\arraystretch}{1.05}
\setlength{\tabcolsep}{6pt}
\small
\begin{tabular}{lccc}
\toprule
\textbf{Metric} & \textbf{Hand-drawn} & \textbf{Bounding Box} & \textbf{Tight Mask} \\
\midrule
CLIP-T$\uparrow$     & \textbf{25.00} & \underline{24.97} & 24.81 \\
L.CLIP-T$\uparrow$   & 21.32 & \underline{21.31} & \textbf{21.59} \\
M.PSNR$\uparrow$     & 28.91 & \underline{29.01} & \textbf{29.53} \\
L.DINO$\uparrow$     & 0.8243 & \underline{0.8269} & \textbf{0.8504} \\
CLIP-F$\uparrow$     & 0.9729 & \underline{0.9734} & \textbf{0.9781} \\
Warp-Err$\downarrow$ & \underline{1.192} & \textbf{1.191} & 1.392 \\
\bottomrule
\end{tabular}
\vspace{-1mm}
\end{table}

\begin{figure*}[t]
    \centering
    \includegraphics[width=\textwidth]{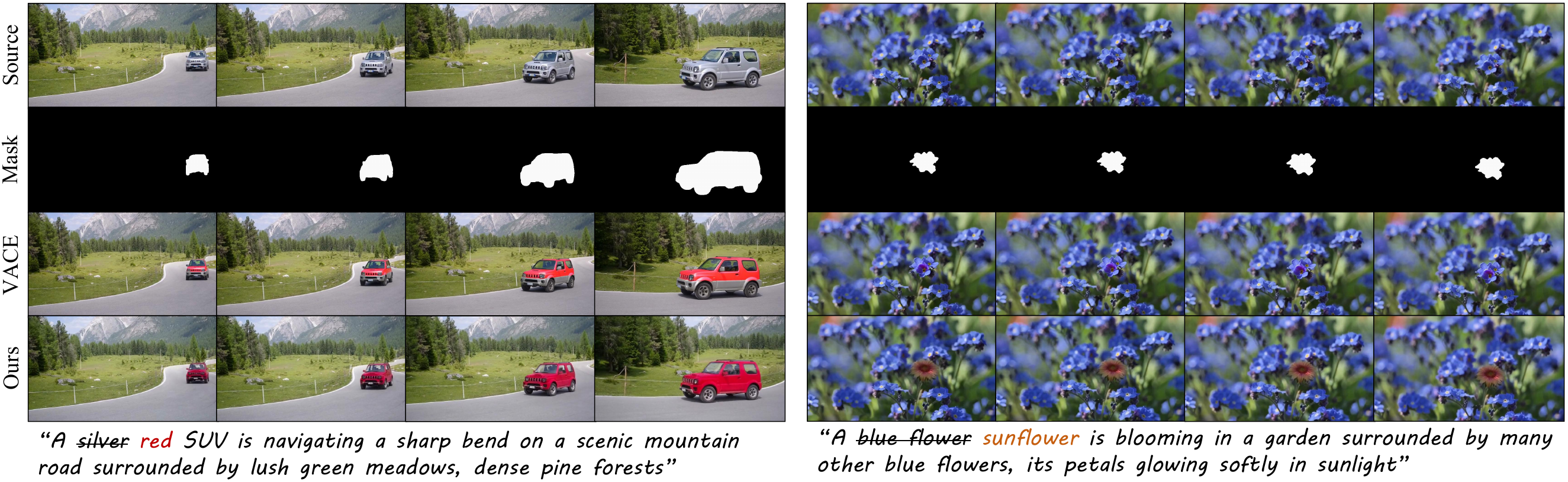}
    \vskip -0.1in
\caption{\textbf{Comparison with the inpainting-based method VACE~\cite{jiang2025vace}.} 
While VACE~\cite{jiang2025vace} is a unified \emph{training-based} framework for mask-based video inpainting, it often suffers from \emph{under-editing} issues, failing to edit the entire masked region or even yielding negligible effects. 
In contrast, our \emph{training-free} FlowAnchor accommodates diverse editing types, delivering highly precise edits that strictly align with the target text while maintaining structural stability.}
    \label{fig:compare_vace}
\end{figure*}

\section{Robustness to Mask Granularity}
\label{subsec:mask}

We further evaluate FlowAnchor on Anchor-Bench using masks of different granularity, including hand-drawn masks, coarse bounding boxes, and tight masks.
The quantitative results are reported in \cref{tab:mask_granularity}.
Overall, FlowAnchor remains effective across all mask forms, showing only moderate variation under coarser masks.
In particular, both hand-drawn masks and bounding boxes still achieve competitive CLIP-T and L.CLIP-T scores, indicating that FlowAnchor does not rely on pixel-accurate masks to establish the target semantics.

Compared with coarse masks, tight masks provide better local fidelity and structure preservation, as reflected by higher L.DINO and M.PSNR.
They also lead to the best CLIP-F score, showing stronger temporal consistency.
We further provide qualitative comparisons in \cref{fig:mask_blue}, where FlowAnchor produces highly consistent editing results across all mask granularities.
This robustness makes FlowAnchor particularly suitable for user-interactive editing scenarios, where the target region is often specified by rough inputs rather than precise segmentation masks.

\section{Additional Comparisons with FlowDirector}
\label{sec:addition}

FlowDirector~\cite{li2025flowdirector} performs spatially constrained editing by deriving an implicit mask from the source and target CA maps and using it to gate the corrected editing flow.
Its update can be written as
\begin{equation}
\hat V_{\mathrm{edit}} = \tilde V_{\mathrm{edit}} \odot \tilde M,
\end{equation}
where \(\tilde M\) is a softened spatial mask constructed from CA responses.
In this way, the editing effect is restricted to regions selected by the attention-derived mask.

\begin{figure*}[t]
    \centering
    \includegraphics[width=\textwidth]{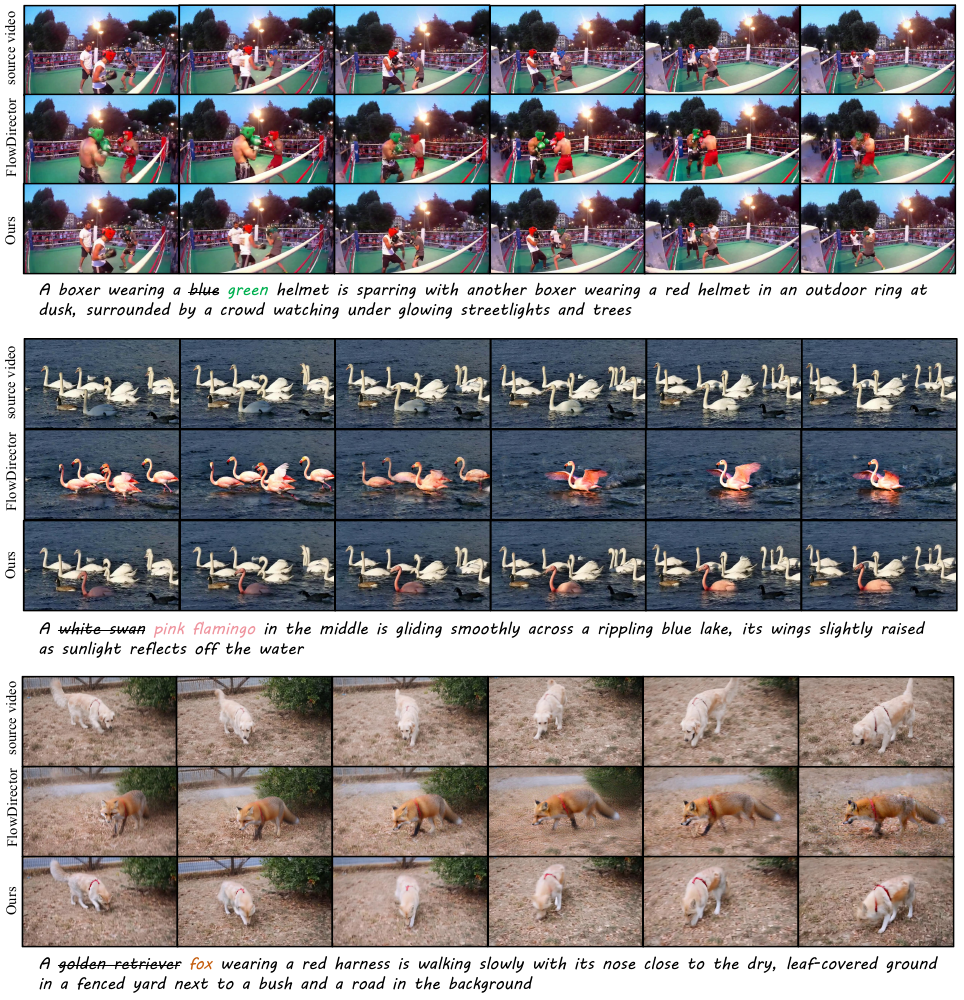}
    \vskip -0.1in
    \caption{\textbf{Comparison with FlowDirector.}
    FlowDirector derives an implicit spatial mask from the source and target CA maps and uses it to gate the corrected editing flow, i.e., $\hat{V}^{\mathrm{edit}}=\tilde{V}^{\mathrm{edit}}\odot \tilde{M}$, where $\tilde{M}$ is a softened mask constructed from CA responses.
    However, such attention-derived masks are often spatially ambiguous and temporally unstable, which leads to editing leakage and corruption of background content.
    This issue becomes more severe in multi-object scenes and under fast motion.
    In contrast, FlowAnchor stabilizes the editing signal itself, yielding more accurate localization, better background preservation, and stronger temporal consistency.}
    \label{fig:compare_flowdirector}
\end{figure*}

However, we find that this design critically depends on the quality of CA localization.
As also observed in Wan-Edit, CA maps are often spatially ambiguous and temporally unstable.
Even in relatively simple scenes, inaccurate attention responses may cause the editing effect to leak into nearby background regions and damage source fidelity.
This issue becomes more severe in multi-object videos or under fast motion, where the target-related responses may drift across different objects or fluctuate over time.
As shown in \cref{fig:compare_flowdirector}, FlowDirector often edits irrelevant regions, corrupts background content, or produces unstable results across frames.

A more fundamental limitation is that FlowDirector directly gates the corrected editing flow using the predicted mask.
As a result, any noise or misalignment in the attention map is immediately propagated into the editing trajectory.
In practice, this tends to preserve or amplify irrelevant responses inside the predicted mask, including background noise, and thus makes the editing behavior highly sensitive to attention errors.

In contrast, FlowAnchor does not treat CA maps as explicit masks for hard spatial gating.
Instead, SAR first refines the attention distribution to improve semantic alignment and localization, and AMM then modulates the editing signal itself in a content-adaptive manner.
Notably, the editing signal
\begin{equation}
\Delta V_{t_i} = V^{\mathrm{tar}}_{t_i} - V^{\mathrm{src}}_{t_i}
\end{equation}
naturally encodes the semantic difference between the target and source conditions, similar in spirit to the correction term discussed in prior flow-based editing methods~\cite{jiao2025uniedit}.
From this perspective, AMM can be interpreted as an adaptive correction mechanism:
\begin{equation}
\Delta V_{t_i}^{\mathrm{AMM}}
=
\Delta V_{t_i}
+
\gamma_F \big(C_{t_i}\odot \Delta V_{t_i}\big),
\end{equation}
where the contrast map \(C_{t_i}\) is derived from the internal signal itself rather than from an external mask.

This distinction is important.
Unlike attention-mask gating, AMM does not uniformly preserve or amplify all responses within a predicted spatial region.
Instead, it strengthens the semantic residuals already encoded in \(\Delta V_{t_i}\).
Therefore, regions with stronger semantic contrast receive larger correction, while weak or irrelevant responses are not blindly amplified.
This greatly reduces the risk of propagating background noise and makes the editing trajectory more stable.

Consequently, compared with FlowDirector, FlowAnchor achieves more accurate localization, better preservation of background and object structure, and stronger temporal consistency across challenging scenarios, especially in multi-object scenes and under fast motion.

\section{Comparison with Inpainting Method}
\label{sec:inpainting}

We further compare FlowAnchor with the representative inpainting-based method VACE~\cite{jiang2025vace}. 
VACE~\cite{jiang2025vace} is a unified \emph{training-based} framework that supports video editing by inpainting the masked region according to the surrounding spatiotemporal context and diverse external conditions. 
Although this design is flexible, we find that it suffers from \emph{under-editing} in text-based localized editing. 
As shown in~\figref{compare_vace}, even when the mask covers the entire car, VACE still edits the object only partially across frames. 
Furthermore, it struggles to perform object replacement, showing negligible editing effects in the ``sunflower'' case. 
In contrast, as a \emph{training-free} approach, our FlowAnchor exhibits high versatility across various editing types. 
It achieves significantly more precise editing that aligns with the target text, while consistently better preserving the original structure and fine-grained appearance details.

\section{Additional Results of FlowAnchor}
\label{sec:more_results}

\begin{figure*}[t]
    \centering
    \includegraphics[width=0.95\textwidth]{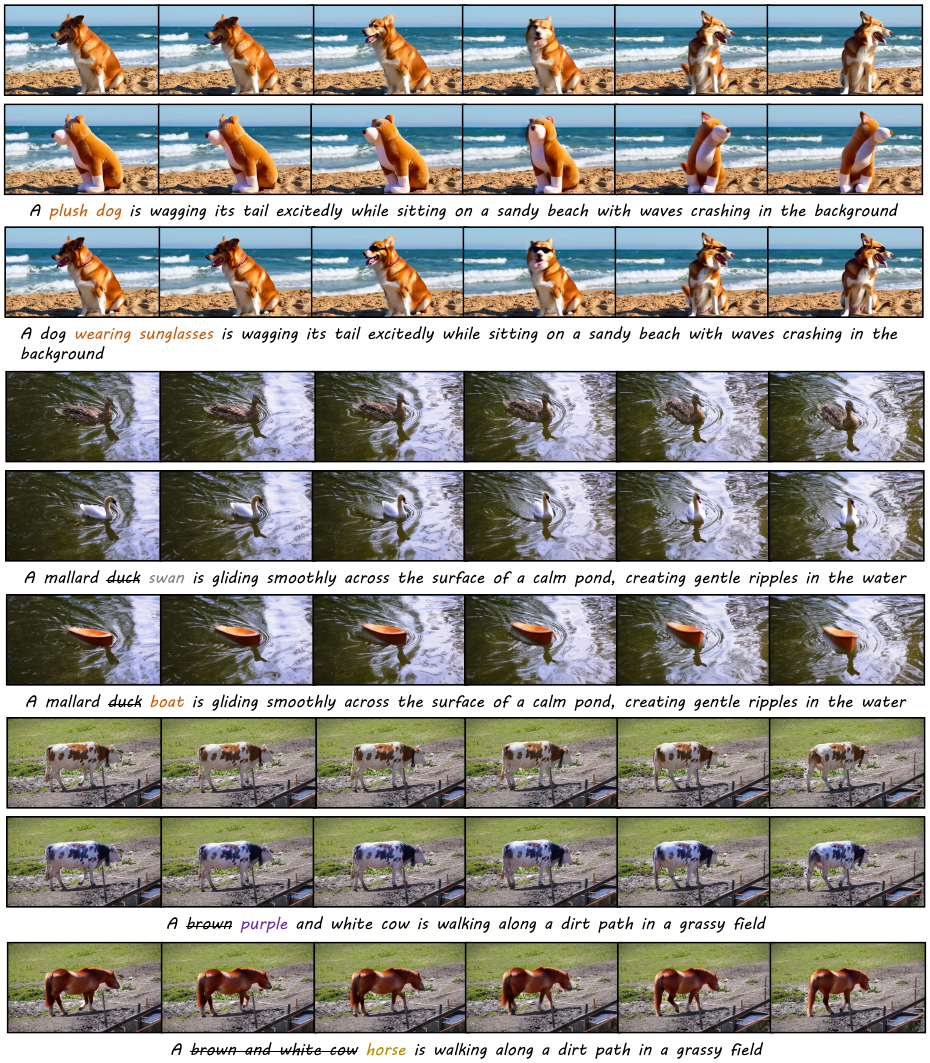}
    \vskip -0.1in
    \caption{\textbf{Qualitative results of FlowAnchor.}
FlowAnchor handles a wide range of editing tasks, including color editing, texture and material modification, object replacement (both rigid and non-rigid), object addition, and localized semantic style transfer.}

    \label{fig:ours_fancy}
\end{figure*}

As shown in Fig.~\ref{fig:ours_fancy}, FlowAnchor supports a diverse range of editing types with high precision and fine-grained control. Specifically, our method enables localized semantic style transfer, as demonstrated by transforming a dog into a plush dog while preserving its structure. It also handles non-rigid object addition, such as adding sunglasses that naturally adapt to the underlying motion and geometry. Furthermore, FlowAnchor supports non-rigid shape editing, as illustrated by transforming a duck into a boat. Importantly, our method achieves delicate appearance editing on fine-grained textures, such as changing the color of the cow’s coat pattern from brown and white to purple and white, while faithfully preserving the original pattern structure. This is particularly challenging, as it requires modifying appearance without disrupting the intricate texture layout.

\begin{figure*}[t]
    \centering
    \includegraphics[width=0.85\textwidth]{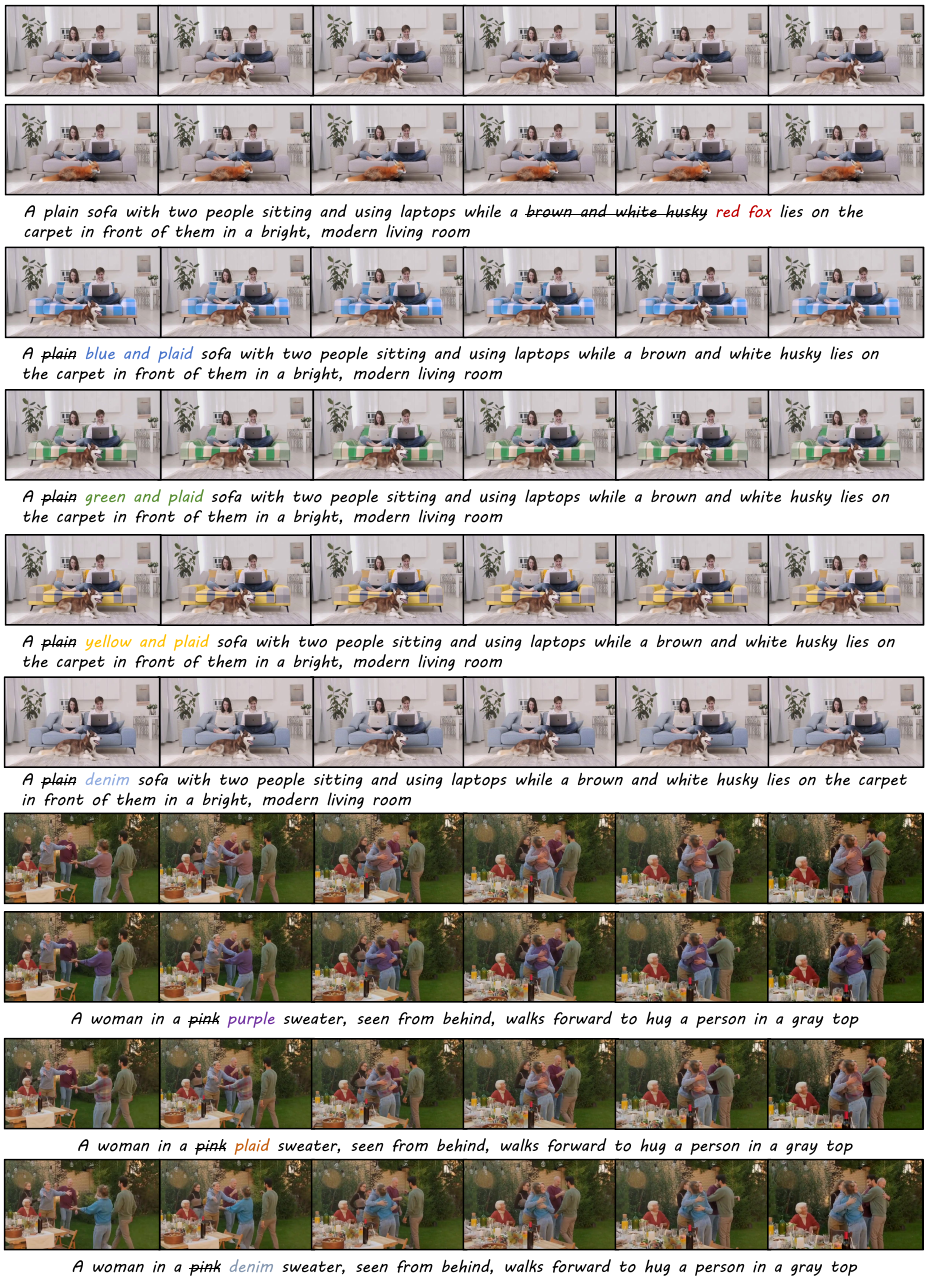}
    \vskip -0.1in
    \caption{\textbf{Qualitative results of FlowAnchor.}
FlowAnchor handles a wide range of editing tasks, including color editing, texture and material modification, object replacement (both rigid and non-rigid), object addition, and localized semantic style transfer.}

    \label{fig:ours_multi}
\end{figure*}

As shown in Fig.~\ref{fig:ours_multi}, FlowAnchor further demonstrates strong robustness in complex multi-object scenarios under diverse editing prompts. Even within a single source video containing multiple interacting objects, our method consistently produces high-quality results across different editing instructions. This highlights its ability to selectively manipulate target regions while preserving the integrity of other objects, while maintaining temporal coherence and structural consistency in cluttered and dynamic scenes.

\begin{figure*}[t]
    \centering
    \includegraphics[width=0.9\textwidth]{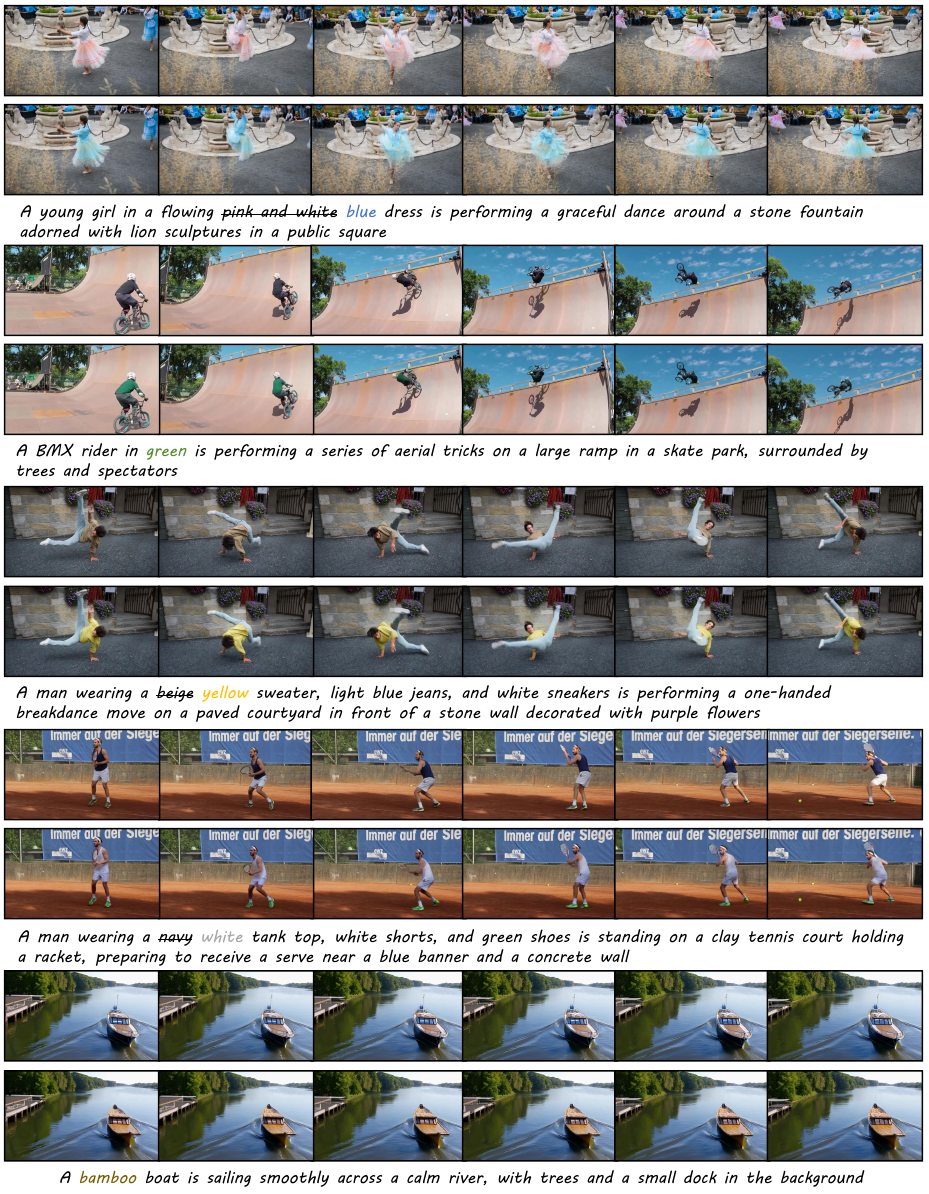}
    \vskip -0.1in
    \caption{\textbf{Qualitative results of FlowAnchor.}
FlowAnchor handles a wide range of editing tasks, including color editing, texture and material modification, object replacement (both rigid and non-rigid), object addition, and localized semantic style transfer.}

    \label{fig:ours_color}
\end{figure*}

FlowAnchor is also effective in challenging video editing scenarios involving fast motion and complex temporal dynamics. As shown in Fig.~\ref{fig:ours_color}, the breakdance example demonstrates that our method can achieve accurate localized editing while preserving temporal coherence under rapid and complicated motion. This example is particularly challenging, as it was previously identified as a failure case in IF-V2V~\cite{kong2025taming}, where I2V-based editing methods struggle with overly complex or fast motions even when additional conditioning is introduced. Notably, their method relies on a large-scale Wan 14B model, whereas FlowAnchor achieves better results using only a 1.3B model. These results highlight the superior robustness and efficiency of FlowAnchor in motion-intensive scenarios.

\begin{figure*}[t]
    \centering
    \includegraphics[width=0.9\textwidth]{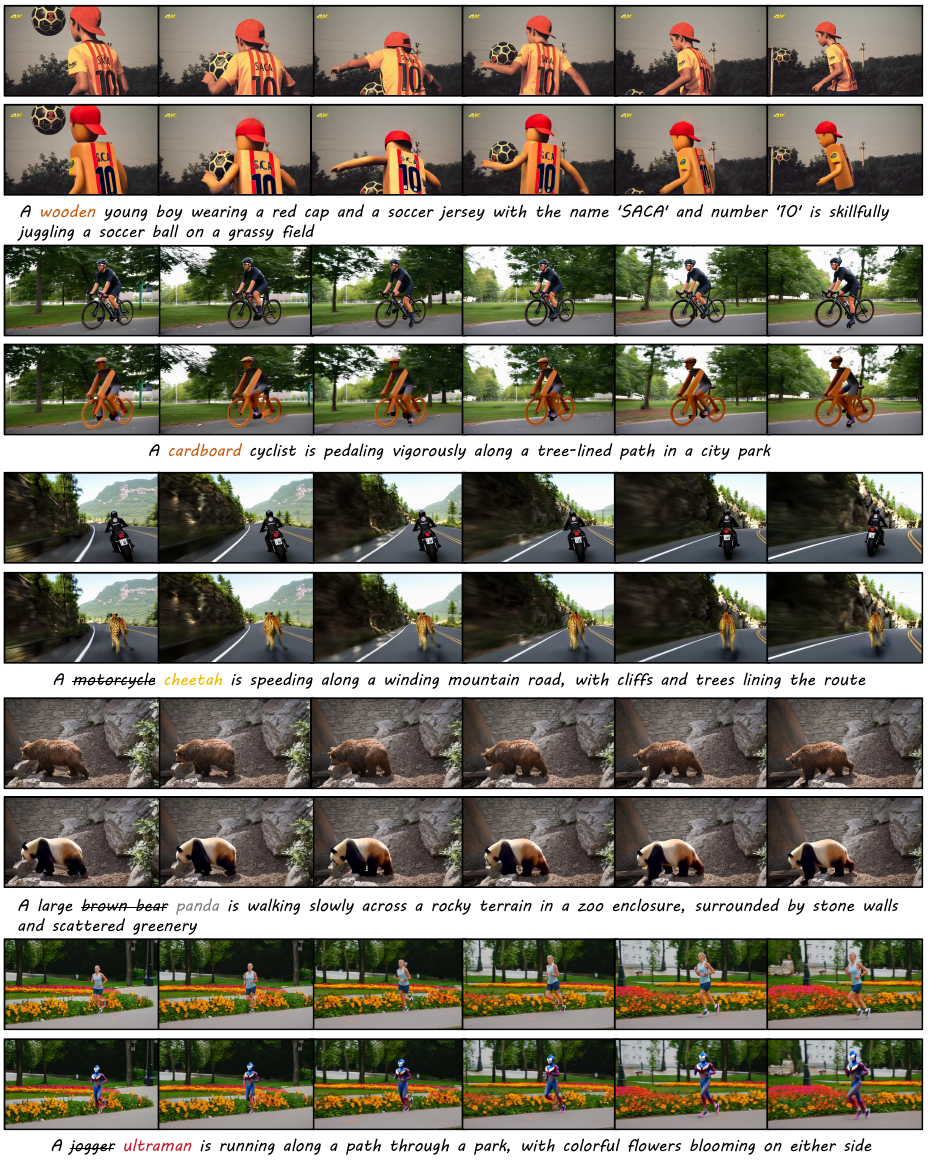}
    \vskip -0.1in
    \caption{\textbf{Qualitative results of FlowAnchor.}
FlowAnchor handles a wide range of editing tasks, including color editing, texture and material modification, object replacement (both rigid and non-rigid), object addition, and localized semantic style transfer.}

    \label{fig:ours_replacement}
\end{figure*}

We further present additional results on FiVE-Bench~\cite{li2025five} in Fig.~\ref{fig:ours_replacement}, demonstrating the effectiveness and generalization ability of FlowAnchor across diverse and challenging benchmark scenarios.


\end{document}